\documentclass{article}

\usepackage[final]{neurips_2024}

\usepackage[utf8]{inputenc} % allow utf-8 input
\usepackage[T1]{fontenc}    % use 8-bit T1 fonts
\usepackage{hyperref}       % hyperlinks
\usepackage{url}            % simple URL typesetting
\usepackage{booktabs}       % professional-quality tables
\usepackage{amsfonts}       % blackboard math symbols
\usepackage{nicefrac}       % compact symbols for 1/2, etc.
\usepackage{microtype}      % microtypography
\usepackage{xcolor}         % colors

\usepackage{mathtools}
\usepackage{subcaption}
\usepackage{wrapfig}
\usepackage{cite}           % magical 
\usepackage[ruled, vlined]{algorithm2e}
\usepackage{sidecap}
\usepackage{array}
\sidecaptionvpos{figure}{c}
\usepackage{tikz}
\usetikzlibrary{arrows,arrows.meta,calc}
\usetikzlibrary{patterns,backgrounds}
\usetikzlibrary{positioning,fit}
\usetikzlibrary{shapes.geometric,shapes.multipart}
\usetikzlibrary{patterns.meta,decorations.pathreplacing,calligraphy}
\usetikzlibrary{tikzmark}
\usetikzlibrary{decorations.pathmorphing}

\newcommand{\x}{\mathbf{x}}

\newcommand{\w}{\mathbf{w}}

\newcommand{\ba}{\mathbf{a}}

\newcommand{\scoref}{\nabla_\x \log p^\tau(\x)}
\newcommand{\scorem}{\mathbf{S}_\theta(\x, \tau)}
\newcommand{\bbe}{\mathbb{E}}
\newcommand{\Tau}{\mathcal{T}}

\newcommand{\wslink}{\href{https://diamond-wm.github.io}{\texttt{https://diamond-wm.github.io}}}

\title{Diffusion for World Modeling:\\Visual Details Matter in Atari\thanks{To prevent confusion, this is the final version of \citep{alonso2023diffusion} and is not related to \citep{ding2024diffusion}.}}

\author{%
  Eloi Alonso\thanks{Equal contribution. $^\ddagger$Equal supervision. Contact: \texttt{eloi.alonso@unige.ch} and \texttt{adam.jelley@ed.ac.uk}}\\
  University of Geneva
  \And
  Adam Jelley$^*$\\ 
  University of Edinburgh
  \And
  Vincent Micheli\\
  University of Geneva
  \And
  Anssi Kanervisto\\
  Microsoft Research
  \And
  Amos Storkey\\
  University of Edinburgh
  \And
  Tim Pearce$^\ddagger$\\\
  Microsoft Research
  \And
  Fran\c{c}ois Fleuret$^\ddagger$\\
  University of Geneva
}
\begin{document}

\maketitle

\begin{abstract}
World models constitute a promising approach for training reinforcement learning agents in a safe and sample-efficient manner. Recent world models predominantly operate on sequences of discrete latent variables to model environment dynamics. However, this compression into a compact discrete representation may ignore visual details that are important for reinforcement learning. Concurrently, diffusion models have become a dominant approach for image generation, challenging well-established methods modeling discrete latents. Motivated by this paradigm shift, we introduce \textsc{diamond} (DIffusion As a Model Of eNvironment Dreams), a reinforcement learning agent trained in a diffusion world model. We analyze the key design choices that are required to make diffusion suitable for world modeling, and demonstrate how improved visual details can lead to improved agent performance. \textsc{diamond} achieves a mean human normalized score of 1.46 on the competitive Atari 100k benchmark; a new best for agents trained entirely within a world model. We further demonstrate that \textsc{diamond}'s diffusion world model can stand alone as an interactive neural game engine by training on static \textit{Counter-Strike: Global Offensive} gameplay. To foster future research on diffusion for world modeling, we release our code, agents, videos and playable world models at \wslink.
\end{abstract}

%  To foster future research on diffusion for world modeling, we release our code, agents and playable world models at \repolink. World model videos are available on the project webpage at: \url{https://diamond-wm.github.io}.
\section{Introduction}
\label{sec:introduction}

Generative models of environments, or ``world models" \citep{ha2018world}, are becoming increasingly important as a component for generalist agents to plan and reason about their environment \citep{lecun2022path}. Reinforcement Learning (RL) has demonstrated a wide variety of successes in recent years \citep{alphago,degrave2022magnetic,ouyang2022training}, but is well-known to be sample inefficient, which limits real-world applications. World models have shown promise for training reinforcement learning agents across diverse environments \citep{hafner2023dreamerv3,muzero2020}, with greatly improved sample-efficiency \citep{ye2021efficientzero}, which can enable learning from experience in the real world \citep{wu2023daydreamer}.

% # Generate noise images 
% from PIL import Image
% import torch
% n = 3
% size = 16
% levels = [0.25, 1.0]
% for i in range(n):
%     for level in levels:
%         x = torch.randn(size, size, 3).div(1.5).clip(-1,1).add(1).div(2)
%         m = torch.rand(size, size, 1) > level
%         m = torch.cat([m] * 3, dim=-1)
%         x[m] = 1
%         Image.fromarray(x.mul(255).byte().numpy()).save(f"noise_{i}_level_{level}.png")

\begin{SCfigure}[][h]
%\begin{wrapfigure}{R}{0.5\linewidth}
% \begin{figure}[t!]

\vspace{-5mm}
\resizebox{0.5\linewidth}{!}{%

\begin{tikzpicture}[
    image/.style={black,fill=white,draw,inner sep=0,minimum width=1.1cm,minimum height=1.1cm},
    maps/.style={->,shorten <=1pt,shorten >=1pt,}
  ]

\def\horigap{0.75cm}
\def\vertgap{0.75cm}
\def\smallvertgap{-0.15cm}

%%%%%%%%%%%%%%%%%%%%%%%%%%%%%%%%%%%%%%%%%%%%
% t=0

\path
node[image] (cond n=0 t=0) at (0, 0) {}
node[image] (cond n=1 t=0) at ($(\tikzlastnode)+(3pt,-2pt)$) {}
node[image] (cond n=2 t=0) at ($(\tikzlastnode)+(3pt,-2pt)$) {}
node[image] (cond n=3 t=0) at ($(\tikzlastnode)+(3pt,-2pt)$) {}
node[image] (cond n=4 t=0) at ($(\tikzlastnode)+(3pt,-2pt)$) {$\x^0_{t-1}$}
node[image,above=\vertgap of \tikzlastnode,xshift=-6pt] (X k=0 t=0) {\includegraphics[width=1.1cm]{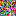}}
node[minimum height=1cm,above=\vertgap of \tikzlastnode] (etc 1 t=0) {$\vdots$}
node[image,above=\smallvertgap of \tikzlastnode] (X k=1 t=0) {\includegraphics[width=1.1cm]{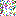}}
node[minimum height=1cm,above=\vertgap of \tikzlastnode] (etc 2 t=0) {$\vdots$}
node[image,above=\smallvertgap of \tikzlastnode] (X k=K t=0) {}
node[right=\horigap of \tikzlastnode,inner sep=2pt,circle,minimum width=1.1cm,draw] (A t=0) {$a_t$}
;

\node[fill=white,inner sep=2pt] at (X k=0 t=0) {$\x^\Tau_t$};
\node[draw=none,fill=white,inner sep=2pt] at (X k=1 t=0) {$\x^\tau_t$};
\node[draw=none,fill=white,inner sep=2pt] at (X k=K t=0) {$\x^0_t$};

%%%%%%%%%%%%%%%%%%%%%%%%%%%%%%%%%%%%%%%%%%%%
% t=1

\path
node[image,right=3cm of {cond n=0 t=0}] (cond n=0 t=1) at (0, 0) {}
node[image] (cond n=1 t=1) at ($(\tikzlastnode)+(3pt,-2pt)$) {}
node[image] (cond n=2 t=1) at ($(\tikzlastnode)+(3pt,-2pt)$) {}
node[image] (cond n=3 t=1) at ($(\tikzlastnode)+(3pt,-2pt)$) {}
node[image] (cond n=4 t=1) at ($(\tikzlastnode)+(3pt,-2pt)$) {$\x^0_t$}
node[image,above=\vertgap of \tikzlastnode,xshift=-6pt] (X k=0 t=1) {\includegraphics[width=1.1cm]{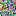}}
node[minimum height=1cm,above=\vertgap of \tikzlastnode] (etc 1 t=1) {$\vdots$}
node[image,above=\smallvertgap of \tikzlastnode] (X k=1 t=1) {\includegraphics[width=1.1cm]{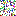}}
node[minimum height=1cm,above=\vertgap of \tikzlastnode] (etc 2 t=1) {$\vdots$}
node[image,above=\smallvertgap of \tikzlastnode] (X k=K t=1) {}
node[right=\horigap of \tikzlastnode,inner sep=2pt,circle,minimum width=1.1cm,draw] (A t=1) {$a_{t+1}$}
;

\node[fill=white,inner sep=2pt] at (X k=0 t=1) {$\x^\Tau_{t+1}$};
\node[draw=none,fill=white,inner sep=2pt] at (X k=1 t=1) {$\x^\tau_{t+1}$};
\node[draw=none,fill=white,inner sep=2pt] at (X k=K t=1) {$\x^0_{t+1}$};

%%%%%%%%%%%%%%%%%%%%%%%%%%%%%%%%%%%%%%%%%%%%
% t=T

\path
node[image,right=7cm of {cond n=0 t=1}] (cond n=0 t=T) at (0, 0) {}
node[image] (cond n=1 t=T) at ($(\tikzlastnode)+(3pt,-2pt)$) {}
node[image] (cond n=2 t=T) at ($(\tikzlastnode)+(3pt,-2pt)$) {}
node[image] (cond n=3 t=T) at ($(\tikzlastnode)+(3pt,-2pt)$) {}
node[image] (cond n=4 t=T) at ($(\tikzlastnode)+(3pt,-2pt)$) {$\x^0_{T-1}$}
node[image,above=\vertgap of \tikzlastnode, xshift=-6pt] (X k=0 t=T) {\includegraphics[width=1.1cm]{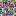}}
node[minimum height=1cm,above=\vertgap of \tikzlastnode] (etc 1 t=T) {$\vdots$}
node[image,above=\smallvertgap of \tikzlastnode] (X k=1 t=T) {\includegraphics[width=1.1cm]{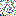}}
node[minimum height=1cm,above=\vertgap of \tikzlastnode] (etc 2 t=T) {$\vdots$}
node[image,above=\smallvertgap of \tikzlastnode] (X k=K t=T) {}
%% node[right=\horigap of \tikzlastnode,inner sep=2pt,circle,minimum width=1.1cm,draw] (A t=T) {$$}
;

\node[fill=white,inner sep=2pt] at (X k=0 t=T) {$\x^\Tau_{T}$};
\node[draw=none,fill=white,inner sep=2pt] at (X k=1 t=T) {$\x^\tau_{T}$};
\node[draw=none,fill=white,inner sep=2pt] at (X k=K t=T) {$\x^0_{T}$};

%%%%%%%%%%%%%%%%%%%%%%%%%%%%%%%%%%%%%%%%%%%%%%%%%%%%%%%%%%%%%%%%%%%%%%

\foreach \t in {0,1}{
  \draw[maps] (X k=K t=\t)--(A t=\t) node[midway,above] {$\pi_\phi$};
}

\foreach \t in {0,1,T}{
%%   \foreach \a/\b in {X k=0 t=\t/etc 1 t=\t, etc 1 t=\t/X k=1 t=\t,X k=1 t=\t/etc 2 t=\t,etc 2 t=\t/X k=K t=\t}{
  \foreach \a/\b in {X k=0 t=\t/etc 1 t=\t,X k=1 t=\t/etc 2 t=\t}{
    \draw[maps] (\a) -- (\b) node[midway,right] {$\mathbf{D}_\theta$};
  }
}

\begin{scope}[on background layer]
  \foreach \n/\a in {n=0 t=0/A,n=1 t=0/A,n=2 t=0/A,n=3 t=0/A,n=4 t=0/$a_{t-1}$,n=0 t=1/A,n=1 t=1/A,n=2 t=1/A,n=3 t=1/A,n=4 t=1/$a_t$,n=0 t=T/A,n=1 t=T/A,n=2 t=T/A,n=3 t=T/A,n=4 t=T/$a_{T-1}$}{
    \node[below=4pt of {cond \n},fill=white,inner sep=2pt,circle,minimum width=1.1cm,draw] (A cond \n) {\a};
  };
\end{scope}

% \node (temp etc) at ($(X k=0 t=1)!0.5!(X k=0 t=T)$) {$\dots$};

\node (temp etc) at ($(X k=K t=1)!0.73!(X k=K t=T)$) {$\dots$};

\draw[->,very thick] ($(X k=K t=0.north west)+(0,5mm)$)--($(X k=K t=T.north east)+(0,5mm)$) node[midway,above,inner sep=5pt] {Environment time ($t$)};

\draw[->,very thick] ($(X k=0 t=0.south west)+(-5mm,0)$)--($(X k=K t=0.north west)+(-5mm,0)$) node[midway,above,inner sep=5pt,rotate=90] (denoising) {Denoising time ($\tau$)};

\draw[draw=none,very thick] ($(X k=0 t=0.south west)+(-5mm,0)$)--++(0,-3) node[midway,above,inner sep=5pt,rotate=90] (conditioning) {Conditioning};

%\node[left=20pt of {cond n=0 t=0},rotate=90,anchor=south] (label) {Conditioning};

%% \node[below=5cm of denoising,rotate=90,yshift=5.5mm] (label) {Conditioning};

%% \draw[->,blue!50,very thick,shorten <=1pt,shorten >=1pt,preaction={draw=white,line width=2pt,-}] (X k=K t=0.south east) to[out=335,in=150] (cond n=4 t=1);

%%%%%%%%%%%%%%%%%%%%%%%%%%%%%%%%%%%%%%%%%%%%

\end{tikzpicture}
}

\caption{Unrolling imagination of \textsc{diamond} over time. The top row depicts a policy $\pi_\phi$ taking a sequence of actions in the imagination of our learned diffusion world model $\mathbf{D}_\theta$. The environment time $t$ flows along the horizontal axis, while the vertical axis represents the denoising time $\tau$ flowing backward from $\Tau$ to $0$. Concretely, given (clean) past observations $\x^0_{<t}$, actions $a_{<t}$, and starting from an initial noisy sample $\x_t^\Tau$, we simulate a reverse noising process $\{\x_t^\tau\}_{\tau =\Tau}^{\tau=0}$ by repeatedly calling $\mathbf{D}_\theta$, and obtain the (clean) next observation $\x_t^0$. The imagination procedure is autoregressive in that the predicted observation $\x_t^0$ and the action $a_t$ taken by the policy become part of the conditioning for the next time step. Animated visualizations of this procedure can be found at \url{https://diamond-wm.github.io}.}

\end{SCfigure}
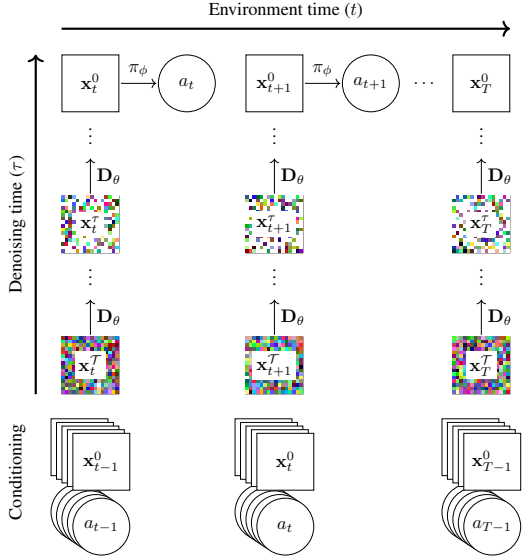
\label{fig:architecture}

Recent world modeling methods \citep{hafner2021mastering,iris2023,robine2023transformer,hafner2023dreamerv3,zhang2023storm} often model environment dynamics as a sequence of discrete latent variables. Discretization of the latent space helps to avoid compounding error over multi-step time horizons. However, this encoding may lose information, resulting in a loss of generality and reconstruction quality. This may be problematic for more real-world scenarios where the information required for the task is less well-defined, such as training autonomous vehicles \citep{hu2023gaia}. In this case, small details in the visual input, such as a traffic light or a pedestrian in the distance, may change the policy of an agent. Increasing the number of discrete latents can mitigate this lossy compression, but comes with an increased computational cost \citep{iris2023}. 

In the meantime, diffusion models \citep{sohl2015difforigin,ho2020DDPM,song_sde} have become a dominant paradigm for high-resolution image generation \citep{ldm_stable_diffusion,podell2023sdxl}. This class of methods, in which the model learns to reverse a noising process, challenges well-established approaches modeling discrete tokens \citep{esser2021taming,ramesh2021zero,muse2023}, and thereby offers a promising alternative to alleviate the need for discretization in world modeling. Additionally, diffusion models are known to be easily conditionable and to flexibly model complex multi-modal distributions without mode collapse. These properties are instrumental to world modeling, since adherence to conditioning should allow a world model to reflect an agent's actions more closely, resulting in more reliable credit assignment, and modeling multi-modal distributions should provide greater diversity of training scenarios for an agent.

Motivated by these characteristics, we propose \textsc{diamond} (DIffusion As a Model Of eNvironment Dreams), a reinforcement learning agent trained in a diffusion world model. 
Careful design choices are necessary to ensure our diffusion world model is efficient and stable over long-time horizons, and we provide a qualitative analysis to illustrate their importance.
\textsc{diamond} achieves a mean human normalized score of 1.46 on the well-established Atari 100k benchmark; a new state of the art for agents trained entirely within a world model.
Additionally, operating in image space has the benefit of enabling our diffusion world model to be a drop-in substitute for the environment, which provides greater insight into world model and agent behaviors. 
In particular, we find the improved performance in some games follows from better modeling of critical visual details. 
To further demonstrate the effectiveness of our world model in isolation, we train \textsc{diamond}'s diffusion world model on $87$ hours of static \textit{Counter-Strike: Global Offensive} (CSGO) gameplay \citep{pearce2022counter} to produce an interactive neural game engine for the popular in-game map, \textit{Dust II}.
We release our code, agents and playable world models at \wslink.

\vspace{-0.1cm}
\section{Preliminaries}
\label{sec:framework}
\vspace{-0.1cm}

\subsection{Reinforcement learning and world models}
\label{subsec:pomdp_and_wm}

We model the environment as a standard Partially Observable Markov Decision Process (\textsc{pomdp}) \citep{sutton2018reinforcement}, $(\mathcal{S}, \mathcal{A}, \mathcal{O},T,R,O,\gamma)$, where $\mathcal{S}$ is a set of states, $\mathcal{A}$ is a set of discrete actions, and $\mathcal{O}$ is a set of image observations. The transition function $T: \mathcal{S} \times \mathcal{A} \times \mathcal{S} \to [0,1]$ describes the environment dynamics $p(\mathbf{s}_{t+1} \mid \mathbf{s}_t, \ba_t)$, and the reward function $R: \mathcal{S} \times \mathcal{A} \times \mathcal{S} \to \mathbb{R}$ maps transitions to scalar rewards. Agents cannot directly access states $s_t$ and only see the environment through image observations $x_t \in \mathcal{O}$, emitted according to observation probabilities $p(\x_t \mid \mathbf{s}_t)$, described by the observation function $O: \mathcal{S} \times \mathcal{O} \to [0,1]$. The goal is to obtain a policy $\pi$ that maps observations to actions in order to maximize the expected discounted return $\mathbb{E}_\pi[\sum_{t \ge 0} \gamma^t r_t]$, where $\gamma \in [0,1]$ is a discount factor. World models \citep{ha2018world} are generative models of environments, i.e. models of $p(s_{t+1},r_{t} \mid s_t, a_t)$. These models can be used as simulated environments to train RL agents \citep{sutton1991dyna} in a sample-efficient manner \citep{wu2023daydreamer}. In this paradigm, the training procedure typically consists of cycling through the three following steps: collect data with the RL agent in the real environment; train the world model on all the collected data; train the RL agent in the world model environment (commonly referred to as "in imagination"). 

\subsection{Score-based diffusion models}
\label{subsec:diffusion}

Diffusion models \citep{sohl2015difforigin} are a class of generative models inspired by non-equilibrium thermodynamics that generate samples by reversing a noising process.

We consider a diffusion process $\{\x^\tau\}_{\tau \in [0,\Tau]}$ indexed by a continuous time variable $\tau \in [0,\Tau]$, with corresponding marginals $\{p^\tau\}_{\tau \in [0,\Tau]}$, and boundary conditions $p^0 = p^{data}$ and $p^\Tau = p^{prior}$, where $p^{prior}$ is a tractable unstructured prior distribution, such as a Gaussian. Note that we use $\tau$ and superscript for the diffusion process time, in order to keep $t$ and subscript for the environment time.

This diffusion process can be described as the solution to a standard stochastic differential equation (SDE) \citep{song_sde},
\begin{equation}
\label{eq:forward_process}
    d\x = \mathbf{f} (\x, \tau) d\tau +  g(\tau) d\w, 
\end{equation}
where $\w$ is the Wiener process (Brownian motion), $\mathbf{f}$ a vector-valued function acting as a drift coefficient, and  $g$ a scalar-valued function known as the diffusion coefficient of the process.

To obtain a generative model, which maps from noise to data, we must reverse this process. Remarkably, \citet{anderson1982reverse} shows that the reverse process is also a diffusion process, running backwards in time, and described by the following SDE,
\begin{equation}
\label{eq:reverse_process}
    d\x = [\mathbf{f} (\x, \tau) - g(\tau)^2 \scoref] d\tau +  g(\tau) d\Bar{\w}, 
\end{equation}
where $\Bar{\w}$ is the reverse-time Wiener process, and $\scoref$ is the (Stein) score function, the gradient of the log-marginals with respect to the support. Therefore, to reverse the forward noising process, we are left to define the functions $f$ and $g$ (in Section \ref{subsec:practical_dwm}), and to estimate the unknown score functions $\scoref$, associated with marginals $\{p^\tau\}_{\tau \in [0,\Tau]}$ along the process. In practice, it is possible to use a single time-dependent score model $\scorem$ to estimate these score functions \citep{song_sde}.

At any point in time, estimating the score function is not trivial since we do not have access to the true score function. Fortunately, \citet{hyvarinen2005estimation} introduces the \textit{score matching} objective, which surprisingly enables training a score model from data samples without the knowledge of the underlying score function. To access samples from the marginal $p^\tau$, we need to simulate the forward process from time $0$ to time $\tau$, as we only have clean data samples. This is costly in general, but if $f$ is affine, we can analytically reach any time $\tau$ in the forward process in a single step, by applying a Gaussian perturbation kernel $p^{0\tau}$ to clean data samples \citep{song_sde}. Since the kernel is differentiable, score matching simplifies to a \textit{denoising score matching} objective \citep{vincent2011connection},

\begin{equation}
\label{eq:denoising_sm}
    \mathcal{L}(\theta) = \bbe \left[ \Vert \mathbf{S}_\theta(\x^\tau, \tau) - \nabla_{\x^\tau} \log p^{0\tau}(\x^\tau \mid \x^0) \Vert^2 \right],
\end{equation}

where the expectation is over diffusion time $\tau$, noised sample $\x^\tau \sim p^{0\tau}(\x^\tau \mid \x^0)$, obtained by applying the $\tau$-level perturbation kernel to a clean sample $\x^0 \sim p^{data}(\x^0)$. Importantly, as the kernel $p^{0\tau}$ is a known Gaussian distribution, this objective becomes a simple $L_2$ reconstruction loss,

\begin{equation}
\label{eq:reconstruction_sm}
     \mathcal{L}(\theta) = \bbe \left[ \Vert \mathbf{D}_\theta(\x^\tau, \tau) - \x^{0} \Vert^2 \right],
\end{equation}

with reparameterization $\mathbf{D}_\theta(\x^\tau, \tau) = \mathbf{S}_\theta(\x^\tau, \tau) \sigma^2(\tau) + \x^\tau$, where $\sigma(\tau)$ is the variance of the $\tau$-level perturbation kernel.

\subsection{Diffusion for world modeling}
\label{subsec:dwm_training}

%The score-based diffusion model described in Section \ref{subsec:diffusion} provides an unconditional generative model of $p_{data}$. To serve as a world model, we need a conditional generative model of the environment dynamics, $p(s_{t+1} \mid s_t, a_t)$. Since we consider the more general case of a \textsc{pomdp} where the Markovian state is unknown, we instead approximate this state with a buffer of recent observations and actions. We condition a diffusion model  with this buffer, to estimate $p(\x_{t+1} \mid \x_{\le t}, a_{\le t})$ and generate the next observation directly, as demonstrated in Figure \ref{fig:architecture}. This modifies Equation \ref{eq:reconstruction_sm} as follows,

The score-based diffusion model described in Section \ref{subsec:diffusion} provides an unconditional generative model of $p_{data}$. To serve as a world model, we need a conditional generative model of the environment dynamics, $p(\x_{t+1} \mid \x_{\le t}, a_{\le t})$, where we consider the general case of a \textsc{pomdp}, in which the Markovian state $s_t$ is unknown and can be approximated from past observations and actions. We can condition a diffusion model on this history, to estimate and generate the next observation directly, as shown in Figure \ref{fig:architecture}. This modifies Equation \ref{eq:reconstruction_sm} as follows,
\begin{equation}
\label{eq:denoising_sm_conditional}
     \mathcal{L}(\theta) = \bbe \left[ \Vert \mathbf{D}_\theta(\x_{t+1}^\tau, \tau, \x_{\le t}^0, a_{\le t}) - \x_{t+1}^0 \Vert^2 \right].
\end{equation}
\vspace{-5mm}

During training, we sample a trajectory segment $\x_{\le t}^0, a_{\le t}, \x_{t+1}^0$ from the agent's replay dataset, and we obtain the noised next observation $\x_{t+1}^\tau \sim p^{0\tau}(\x_{t+1}^\tau \mid \x_{t+1}^0)$ by applying the $\tau$-level perturbation kernel. In summary, this diffusion process for world modeling resembles the standard diffusion process described in Section \ref{subsec:diffusion}, with a score model conditioned on past observations and actions.

To sample the next observation, we iteratively solve the reverse SDE in Equation \ref{eq:reverse_process}, as illustrated in Figure \ref{fig:architecture}. While we can in principle resort to any ODE or SDE solver, there is an inherent trade-off between sampling quality and Number of Function Evaluations (NFE), that directly determines the inference cost of the diffusion world model (see Appendix \ref{appendix:sampling} for more details).

\section{Method}
\label{sec:method}

\subsection{Practical choice of diffusion paradigm}
\label{subsec:practical_dwm}

Building on the background provided in Section \ref{sec:framework}, we now introduce \textsc{diamond} as a practical realization of a diffusion-based world model. In particular, we now define the drift and diffusion coefficients $\mathbf{f}$ and $g$ introduced in Section \ref{subsec:diffusion}, corresponding to a particular choice of diffusion paradigm. While \textsc{ddpm} \citep{ho2020DDPM} is an example of one such choice (as described in Appendix \ref{app:ddpm}) and would historically be the natural candidate, we instead build upon the \textsc{edm} formulation proposed in \citet{karras2022elucidating}. The practical implications of this choice are discussed in Section \ref{subsec:diffusion_choice}. In what follows, we describe how we adapt \textsc{edm} to build our diffusion-based world model.

We consider the perturbation kernel $p^{0\tau}(\x_{t+1}^\tau \mid \x_{t+1}^0) = \mathcal{N}(\x_{t+1}^\tau; \x_{t+1}^0, \sigma^2(\tau) \mathbf{I})$, where $\sigma(\tau)$ is a real-valued function of diffusion time called the noise schedule. This corresponds to setting the drift and diffusion coefficients to $\mathbf{f}(\x, \tau) = \mathbf{0}$ (affine) and $g(\tau) = \sqrt{2 \dot \sigma(\tau) \sigma(\tau)}$.

We use the network preconditioning introduced by \citet{karras2022elucidating} and so parameterize $\mathbf{D}_\theta$ in Equation \ref{eq:denoising_sm_conditional} as the weighted sum of the noised observation and the prediction of a neural network $\mathbf{F}_\theta$,
\begin{equation}
\label{eq:karras_wrappers} 
    \mathbf{D}_\theta(\x_{t+1}^\tau, y_t^\tau) = c_\text{skip}^\tau \; \x_{t+1}^\tau + c_\text{out}^\tau \; \mathbf{F}_\theta \big( c_\text{in}^\tau \; \x_{t+1}^\tau, y_t^\tau \big),
\end{equation}
where for brevity we define $y_t^\tau \coloneqq (c_\text{noise}^\tau, \x^0_{\le t}, a_{\le t})$ to include all conditioning variables.

The preconditioners $c_\text{in}^\tau$ and $c_\text{out}^\tau$ are selected to keep the network's input and output at unit variance for any noise level $\sigma(\tau)$, $c_\text{noise}^\tau$ is an empirical transformation of the noise level, and $c_\text{skip}^\tau$ is given in terms of $\sigma(\tau)$ and the standard deviation of the data distribution $\sigma_\text{data}$, as $c_{skip}^\tau = \sigma_{data}^2/(\sigma_{data}^2 + \sigma^2(\tau))$. These preconditioners are fully described in Appendix \ref{appendix:karras_conditioners}.

Combining Equations \ref{eq:denoising_sm_conditional} and \ref{eq:karras_wrappers} provides insight into the training objective of $\mathbf{F}_\theta$,
\begin{align}
\label{eq:effective_obj}
\mathcal{L}(\theta)  = \bbe \Big[ \Vert 
\underbrace{\mathbf{F}_\theta \big( c_\text{in}^\tau \x_{t+1}^\tau, y_t^\tau \big)}_\text{Network prediction} - 
\underbrace{\frac{1}{c_\text{out}^\tau} \big( \x_{t+1}^0 - c_\text{skip}^\tau \x_{t+1}^\tau\big)}_\text{Network training target}
\Vert^2 \Big].
\end{align}
The network training target adaptively mixes signal and noise depending on the degradation level $\sigma(\tau)$.
When $\sigma(\tau) \gg \sigma_\text{data}$, we have $c_\text{skip}^\tau \to 0$, and the training target for $\mathbf{F}_\theta$ is dominated by the clean signal $\x_{t+1}^0$. Conversely, when the noise level is low, $\sigma(\tau) \to 0$, we have $c_\text{skip}^\tau \to 1$, and the target becomes the difference between the clean and the perturbed signal, i.e. the added Gaussian noise. Intuitively, this prevents the training objective to become trivial in the low-noise regime. In practice, this objective is high variance at the extremes of the noise schedule, so \citet{karras2022elucidating} sample the noise level $\sigma(\tau)$ from an empirically chosen log-normal distribution in order to concentrate the training around medium-noise regions, as described in Appendix \ref{appendix:karras_conditioners}.

We use a standard U-Net 2D for the vector field $\mathbf{F}_\theta$ \citep{ronneberger2015unet}, and we keep a buffer of $L$ past observations and actions that we use to condition the model. We concatenate these past observations to the next noisy observation channel-wise, and we input actions through adaptive group normalization layers \citep{adagn} in the residual blocks \citep{He2015} of the U-Net.

As discussed in Section \ref{subsec:dwm_training} and Appendix \ref{appendix:sampling}, there are many possible sampling methods to generate the next observation from the trained diffusion model. While our codebase supports a variety of sampling schemes, we found Euler's method to be effective without incurring the cost of additional NFE required by higher order samplers, or the unnecessary complexity of stochastic sampling.

\subsection{Reinforcement learning in imagination}
\label{subsec:rl}

Given the diffusion model from Section \ref{subsec:practical_dwm}, we now complete our world model with a reward and termination model, required for training an RL agent in imagination. Since estimating the reward and termination are scalar prediction problems, we use a separate model $R_\psi$ consisting of standard \textsc{cnn} \citep{cnn_lecun,He2015} and \textsc{lstm} \citep{lstm,Gers2000} layers to handle partial observability. The RL agent involves an actor-critic network parameterized by a shared \textsc{cnn-lstm} with policy and value heads. The policy $\pi_\phi$ is trained with \textsc{reinforce} with a value baseline, and we use a Bellman error with $\lambda$-returns to train the value network $V_\phi$, similar to \citet{iris2023}. We train the agent entirely in imagination as described in Section \ref{subsec:pomdp_and_wm}. The agent only interacts with the real environment for data collection. After each collection stage, the current world model is updated by training on all data collected so far. Then, the agent is trained with RL in the updated world model environment, and these steps are repeated. This procedure is detailed in Algorithm \ref{alg:diamond}, and is similar to \citet{kaiser2019atari100k,hafner2020dream,iris2023}. We provide architecture details, hyperparameters, and RL objectives in Appendices \ref{app:architectures}, \ref{app:hyperparams}, \ref{appendix:rl_actor_critic}, respectively.

\vspace{-5mm}
\section{Experiments}
\label{sec:experiments}

\vspace{-1mm}
\subsection{Atari 100k benchmark}
\label{subsec:atari100k}
For comprehensive evaluation of \textsc{diamond} we use the established Atari 100k benchmark \citep{kaiser2019atari100k}, consisting of 26 games that test a wide range of agent capabilities. 
%These environments are visually diverse, with performance in many games depending on differences in the shapes, sizes and colors of objects. 
For each game, an agent is only allowed to take 100k actions in the environment, which is roughly equivalent to 2 hours of human gameplay, to learn to play the game before evaluation. 
As a reference, unconstrained Atari agents are usually trained for 50 million steps, a 500 fold increase in experience. 
We trained \textsc{diamond} from scratch for 5 random seeds on each game. 
Each run utilized around 12GB of VRAM and took approximately 2.9 days on a single Nvidia RTX 4090 (1.03 GPU years in total).

\begin{table*}[h]
    \caption{Returns on the 26 games of the Atari 100k benchmark after 2 hours of real-time experience, and human-normalized aggregate metrics. Bold numbers indicate the best performing methods. \textsc{diamond} notably outperforms other world model baselines in terms of mean score over 5 seeds.}
    \label{tab:atari_results_full}
\vspace{-3mm}
\begin{center}
\begin{small}
\centering
\scalebox{0.86}{
\centering

\begin{tabular}{lrr rrrrrr}
\toprule
%\multicolumn{3}{c}{} & \multicolumn{2}{c}{Model-free} & \multicolumn{5}{c}{Imagination-based} \\
%\cmidrule(lr){4-5} \cmidrule(lr){6-11}

%%%%%%%%%%%%%%%%%%%%%%%%%%%%%%%%%%%%%%%%%%%%%%%%%%%%%%%%%%%%%%%%%%%%
Game                 &  Random    &  Human     &  SimPLe    &  TWM                &  IRIS              &  DreamerV3          &  STORM              &  \textsc{diamond} (ours)  \\
\midrule
Alien                &  227.8     &  7127.7    &  616.9     &  674.6              &  420.0             &  959.0              &  \textbf{983.6}     &  744.1                    \\
Amidar               &  5.8       &  1719.5    &  74.3      &  121.8              &  143.0             &  139.0              &  204.8              &  \textbf{225.8}           \\
Assault              &  222.4     &  742.0     &  527.2     &  682.6              &  1524.4            &  706.0              &  801.0              &  \textbf{1526.4}          \\
Asterix              &  210.0     &  8503.3    &  1128.3    &  1116.6             &  853.6             &  932.0              &  1028.0             &  \textbf{3698.5}          \\
BankHeist            &  14.2      &  753.1     &  34.2      &  466.7              &  53.1              &  \textbf{649.0}     &  641.2              &  19.7                     \\
BattleZone           &  2360.0    &  37187.5   &  4031.2    &  5068.0             &  13074.0           &  12250.0            &  \textbf{13540.0}   &  4702.0                   \\
Boxing               &  0.1       &  12.1      &  7.8       &  77.5               &  70.1              &  78.0               &  79.7               &  \textbf{86.9}            \\
Breakout             &  1.7       &  30.5      &  16.4      &  20.0               &  83.7              &  31.0               &  15.9               &  \textbf{132.5}           \\
ChopperCommand       &  811.0     &  7387.8    &  979.4     &  1697.4             &  1565.0            &  420.0              &  \textbf{1888.0}    &  1369.8                   \\
CrazyClimber         &  10780.5   &  35829.4   &  62583.6   &  71820.4            &  59324.2           &  97190.0            &  66776.0            &  \textbf{99167.8}         \\
DemonAttack          &  152.1     &  1971.0    &  208.1     &  350.2              &  \textbf{2034.4}   &  303.0              &  164.6              &  288.1                    \\
Freeway              &  0.0       &  29.6      &  16.7      &  24.3               &  31.1              &  0.0                &  \textbf{33.5}      &  33.3                     \\
Frostbite            &  65.2      &  4334.7    &  236.9     &  \textbf{1475.6}    &  259.1             &  909.0              &  1316.0             &  274.1                    \\
Gopher               &  257.6     &  2412.5    &  596.8     &  1674.8             &  2236.1            &  3730.0             &  \textbf{8239.6}    &  5897.9                   \\
Hero                 &  1027.0    &  30826.4   &  2656.6    &  7254.0             &  7037.4            &  \textbf{11161.0}   &  11044.3            &  5621.8                   \\
Jamesbond            &  29.0      &  302.8     &  100.5     &  362.4              &  462.7             &  445.0              &  \textbf{509.0}     &  427.4                    \\
Kangaroo             &  52.0      &  3035.0    &  51.2      &  1240.0             &  838.2             &  4098.0             &  4208.0             &  \textbf{5382.2}          \\
Krull                &  1598.0    &  2665.5    &  2204.8    &  6349.2             &  6616.4            &  7782.0             &  8412.6             &  \textbf{8610.1}          \\
KungFuMaster         &  258.5     &  22736.3   &  14862.5   &  24554.6            &  21759.8           &  21420.0            &  \textbf{26182.0}   &  18713.6                  \\
MsPacman             &  307.3     &  6951.6    &  1480.0    &  1588.4             &  999.1             &  1327.0             &  \textbf{2673.5}    &  1958.2                   \\
Pong                 &  -20.7     &  14.6      &  12.8      &  18.8               &  14.6              &  18.0               &  11.3               &  \textbf{20.4}            \\
PrivateEye           &  24.9      &  69571.3   &  35.0      &  86.6               &  100.0             &  882.0              &  \textbf{7781.0}    &  114.3                    \\
Qbert                &  163.9     &  13455.0   &  1288.8    &  3330.8             &  745.7             &  3405.0             &  \textbf{4522.5}    &  4499.3                   \\
RoadRunner           &  11.5      &  7845.0    &  5640.6    &  9109.0             &  9614.6            &  15565.0            &  17564.0            &  \textbf{20673.2}         \\
Seaquest             &  68.4      &  42054.7   &  683.3     &  \textbf{774.4}     &  661.3             &  618.0              &  525.2              &  551.2                    \\
UpNDown              &  533.4     &  11693.2   &  3350.3    &  \textbf{15981.7}   &  3546.2            &  9234.0             &  7985.0             &  3856.3                   \\
\midrule
\#Superhuman (↑)     &  0         &  N/A       &  1         &  8                  &  10                &  9                  &  10                 &  \textbf{11}              \\
Mean (↑)             &  0.000     &  1.000     &  0.332     &  0.956              &  1.046             &  1.097              &  1.266              &  \textbf{1.459}           \\
IQM (↑)              &  0.000     &  1.000     &  0.130     &  0.459              &  0.501             &  0.497              &  \textbf{0.636}     &  \textbf{0.641}           \\
%%%%%%%%%%%%%%%%%%%%%%%%%%%%%%%%%%%%%%%%%%%%%%%%%%%%%%%%%%%%%%%%%%

\bottomrule
\end{tabular}
 }
\end{small}
\end{center}
\vspace{-7mm}
\end{table*}

We compare with other recent methods training an agent entirely within a world model in Table \ref{tab:atari_results_full}, including \textsc{storm} \citep{zhang2023storm}, DreamerV3 \citep{hafner2023dreamerv3}, \textsc{iris} \citep{iris2023}, \textsc{twm} \citep{robine2023transformer}, and SimPle \citep{kaiser2019atari100k}. A broader comparison to model-free and search-based methods, including \textsc{bbf} \citep{schwarzer2023bigger} and EfficientZero \citep{ye2021efficientzero}, the current best performing methods on this benchmark, is provided in Appendix \ref{app:additonal_baselines}. \textsc{bbf} and EfficientZero use techniques that are orthogonal and not directly comparable to our approach, such as using periodic network resets in combination with hyperparameter scheduling for \textsc{bbf}, and computationally expensive lookahead Monte-Carlo tree search for EfficientZero. Combining these additional components with our world model would be an interesting direction for future work.

\subsection{Results on the Atari 100k benchmark}
\label{subsec:results}

%%%%%%%%%%%%%%%%%%%%%%%
\begin{wrapfigure}[]{R}{0.50\linewidth}
\vspace{-15mm}
\begin{center}
\centerline{\includegraphics[width=\linewidth]{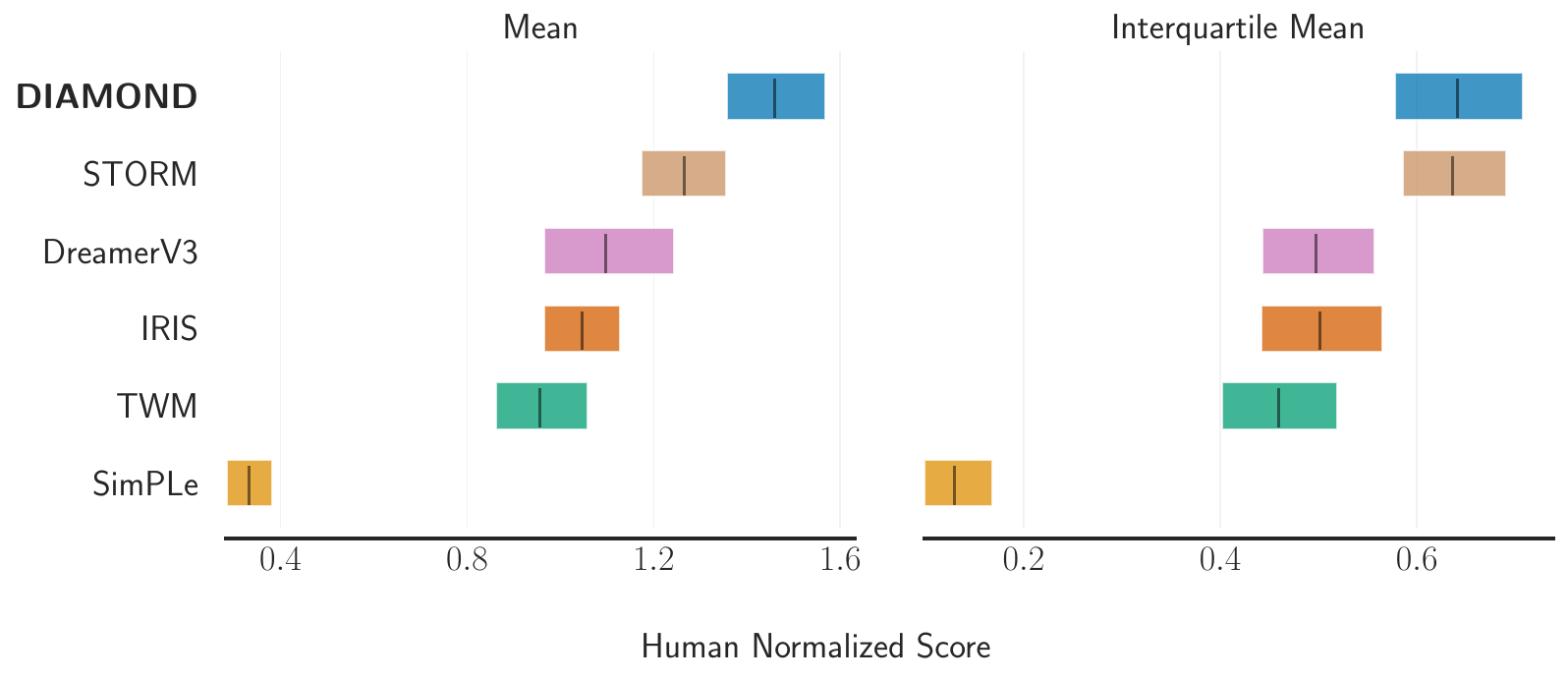}}
\caption{Mean and interquartile mean human normalized scores. \textsc{diamond}, in blue, obtains a mean HNS of 1.46 and an IQM of 0.64.}
\label{fig:results_mean_IQM}
\end{center}
\vskip -15mm
\end{wrapfigure}
%%%%%%%%%%%%%%%%%%%%%%%

Table \ref{tab:atari_results_full} provides scores for all games, and the mean and interquartile mean (IQM) of human-normalized scores (HNS) \citep{dueling_networks}. Following the recommendations of \citet{agarwal2021deep} on the limitations of point estimates, we provide stratified bootstrap confidence intervals for the mean and IQM in Figure \ref{fig:results_mean_IQM}, as well as performance profiles and additional metrics in Appendix \ref{app:performance_profile}.

Our results demonstrate that \textsc{diamond} performs strongly across the benchmark, outperforming human players on 11 games, and achieving a superhuman mean HNS of 1.46, a new best among agents trained entirely within a world model. \textsc{diamond} also achieves an IQM on par with \textsc{storm}, and greater than all other baselines. We find that \textsc{diamond} performs particularly well on environments where capturing small details is important, such as \textit{Asterix}, \textit{Breakout} and \textit{Road Runner}. We provide further qualitative analysis of the visual quality of the world model in Section \ref{subsec:comparison_transformers}. 

\vspace{-2mm}
\section{Analysis}
\label{sec:analysis}
\vspace{-1mm}

\subsection{Choice of diffusion framework}
\label{subsec:diffusion_choice}
\vspace{-1mm}

As explained in Section \ref{sec:framework}, we could in principle use any diffusion model variant in our world model. While \textsc{diamond} utilizes \textsc{edm} \citep{karras2022elucidating} as described in Section \ref{sec:method}, \textsc{ddpm} \citep{ho2020DDPM} would also be a natural candidate, having been used in many image generation applications \citep{ldm_stable_diffusion, ddpm++}. We justify this design decision in this section.

To provide a fair comparison of \textsc{ddpm} with our \textsc{edm} implementation, we train both variants with the same network architecture, on a shared static dataset of 100k frames collected with an expert policy on the game \textit{Breakout}. As discussed in Section \ref{subsec:dwm_training}, the number of denoising steps is directly related to the inference cost of the world model, and so fewer steps will reduce the cost of training an agent on imagined trajectories. \citet{ho2020DDPM} use a thousand denoising steps, and \citet{ldm_stable_diffusion} employ hundreds steps for Stable Diffusion. However, for our world model to be computationally comparable with other world model baselines (such as \textsc{iris} which requires 16 NFE for each timestep), we need at most tens of denoising steps, and preferably fewer. Unfortunately, if the number of denoising steps is set too low, the visual quality will degrade, leading to compounding error. 

\begin{figure}[h]
  \begin{subfigure}{0.49\linewidth}
    \includegraphics[width=\linewidth]{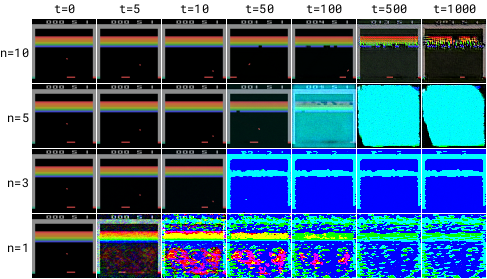}
    \caption{\textsc{ddpm}-based world model trajectories.} \label{fig:denoising_with_ddpm}
  \end{subfigure}%
  \hspace*{\fill}   % maximize separation between the subfigures
  \begin{subfigure}{0.49\textwidth}
    \includegraphics[width=\linewidth]{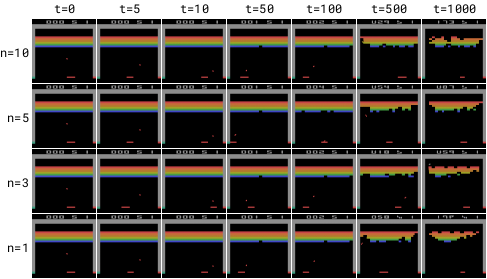}
    \caption{\textsc{edm}-based world model trajectories.} \label{fig:denoising_with_karras}
  \end{subfigure}%

\caption{Imagined trajectories with diffusion world models based on \textsc{ddpm} (left) and \textsc{edm} (right). The initial observation at $t=0$ is common, and each row corresponds to a decreasing number of denoising steps $n$. We observe that \textsc{ddpm}-based generation suffers from compounding error, and that the smaller the number of denoising steps, the faster the error accumulates. In contrast, our \textsc{edm}-based world model appears much more stable, even for $n=1$.\vspace{-5mm}}
\label{fig:denoising_trajectories} 
\end{figure}

To investigate the stability of the diffusion variants, we display imagined trajectories generated autoregressively up to $t=1000$ timesteps in Figure \ref{fig:denoising_trajectories}, for different numbers of denoising steps $n \le 10$. We see that using \textsc{ddpm} (Figure \ref{fig:denoising_with_ddpm}) in this regime leads to severe compounding error, causing the world model to quickly drift out of distribution. In contrast, the \textsc{edm}-based diffusion world model (Figure \ref{fig:denoising_with_karras}) appears much more stable over long time horizons, even for a single denoising step. A quantitative analysis of this compounding error is provided in Appendix~\ref{app:ddpm_drift}.

% %%%%%%%%%%%%%%%%%%%%%%%
% %%%%%%%%%%%%%%%%%%%%%%%

This surprising result is a consequence of the improved training objective described in Equation~\ref{eq:effective_obj}, compared to the simpler noise prediction objective employed by \textsc{ddpm}. While predicting the noise works well for intermediate noise levels, this objective causes the model to learn the identity function when the noise is dominant ($\sigma_{noise}\gg\sigma_{data} \implies \xi_\theta(\x^\tau_{t+1}, y_t^\tau)\to\mathbf{x}^\tau_{t+1}$), where $\xi_\theta$ is the noise prediction network of \textsc{ddpm}. This gives a poor estimate of the score function at the beginning of the sampling procedure, which degrades the generation quality and leads to compounding error.

In contrast, the adaptive mixing of signal and noise employed by \textsc{edm}, described in Section \ref{subsec:practical_dwm}, means that the model is trained to predict the clean image when the noise is dominant ($\sigma_{noise}\gg\sigma_{data} \implies \mathbf{F_\theta}(\mathbf{x}^\tau_{t+1},y_t^\tau)\to\mathbf{x}^0_{t+1}$). This gives a better estimate of the score function in the absence of signal, so the model is able to produce higher quality generations with fewer denoising steps, as illustrated in Figure \ref{fig:denoising_with_karras}.

\subsection{Choice of the number of denoising steps}\label{subsec:denoising_steps}

 While we found that our \textsc{edm}-based world model was very stable with just a single denoising step, as shown for \textit{Breakout} in the last row of Figure \ref{fig:denoising_with_karras}, we discuss here how this choice would limit the visual quality of the model in some cases. We provide more a quantitative analysis in Appendix~\ref{app:denoising_ablation}.

\begin{figure}[h!]
%\vspace{-6mm}
\begin{center}
\includegraphics[width=.64\linewidth]{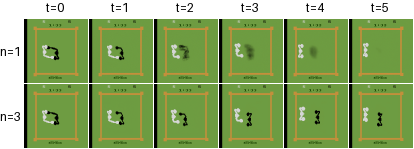}
\caption{Single-step (top row) versus multi-step (bottom row) sampling in \textit{Boxing}. Movements of the black player are unpredictable, so that single-step denoising interpolates between possible outcomes and results in blurry predictions. In contrast, multi-step sampling produces a crisp image by driving the generation towards a particular mode. Interestingly, the policy controls the white player, so his actions are known to the world model. This information removes any ambiguity, and so we observe that both single-step and multi-step sampling correctly predict the white player's position.}
\label{fig:too_optimisitic_4_sure}
\end{center}
\vspace{-4mm}
\end{figure}

As discussed in Section \ref{subsec:diffusion}, our score model is equivalent to a denoising autoencoder \citep{vincent2008extracting} trained with an $L_2$ reconstruction loss. The optimal single-step prediction is thus the expectation over possible reconstructions for a given noisy input, which can be out of distribution if this posterior distribution is multimodal. While some games like \textit{Breakout} have deterministic transitions that can be accurately modeled with a single denoising step (see Figure \ref{fig:denoising_with_karras}), in some other games partial observability gives rise to multimodal observation distributions. In this case, an iterative solver is necessary to drive the sampling procedure towards a particular mode, as illustrated in the game \textit{Boxing} in Figure \ref{fig:too_optimisitic_4_sure}. As a result, we therefore set $n=3$ in all of our experiments.
\vspace{-2mm}
%%%%%%%%%%%%%%%%%%%%%%%
%%%%%%%%%%%%%%%%%%%%%%%

\subsection{Qualitative visual comparison with \textsc{iris}}
\label{subsec:comparison_transformers}

We now compare to \textsc{iris} \citep{iris2023}, a well-established world model that uses a discrete autoencoder \citep{vqvae} to convert images to discrete tokens, and composes these tokens over time with an autoregressive transformer \citep{radford2019language}. For fair comparison, we train both world models on the same static datasets of 100k frames collected with expert policies. This comparison is displayed in Figure \ref{fig:iris_vs_diamond} below.

% %%%%%%%%%%%%%%%%%%%%%%%
\begin{figure}[h]
  \begin{subfigure}{0.49\linewidth}
    \includegraphics[width=\linewidth]{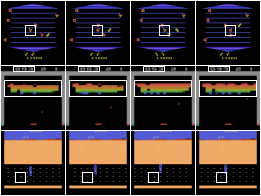}
    \caption{\textsc{iris}} \label{fig:iris_vs_diamond__iris}
  \end{subfigure}%
  \hspace*{\fill}   % maximize separation between the subfigures
  \begin{subfigure}{0.49\textwidth}
    \includegraphics[width=\linewidth]{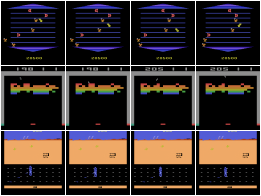}
    \caption{\textsc{diamond}} \label{fig:iris_vs_diamond__diamond}
  \end{subfigure}%
\caption{Consecutive frames imagined with \textsc{iris} (left) and \textsc{diamond} (right). The white boxes highlight inconsistencies between frames, which we see only arise in trajectories generated with \textsc{iris}. In \textit{Asterix} (top row), an enemy (orange) becomes a reward (red) in the second frame, before reverting to an enemy in the third, and again to a reward in the fourth. In \textit{Breakout} (middle row), the bricks and score are inconsistent between frames. In \textit{Road Runner} (bottom row), the rewards (small blue dots on the road) are inconsistently rendered between frames. None of these inconsistencies occur with \textsc{diamond}. In \textit{Breakout}, the score is even reliably updated by +7 when a red brick is broken\protect\footnotemark.}
\label{fig:iris_vs_diamond}
\end{figure}

\footnotetext{\href{https://en.wikipedia.org/wiki/Breakout_(video_game)\#Gameplay}{\texttt{https://en.wikipedia.org/wiki/Breakout\_(video\_game)\#Gameplay}}}
% %%%%%%%%%%%%%%%%%%%%%%%

We see in Figure \ref{fig:iris_vs_diamond} that the trajectories imagined by \textsc{diamond} are generally of higher visual quality and more faithful to the true environment compared to the trajectories imagined by \textsc{iris}. In particular, the trajectories generated by \textsc{iris} contain visual inconsistencies between frames (highlighted by white boxes), such as enemies being displayed as rewards and vice-versa. These inconsistencies may only represent a few pixels in the generated images, but can have significant consequences for reinforcement learning. For example, since an agent should generally target rewards and avoid enemies, these small visual discrepancies can make it more challenging to learn an optimal policy.

These improvements in the consistency of visual details are generally reflected by greater agent performance on these games, as shown in Table \ref{tab:atari_results_full}. Since the agent component of these methods is similar, this improvement can likely be attributed to the world model. 

Finally, we note that this improvement is not simply the result of increased computation. Both world models are rendering frames at the same resolution ($64\times64$), and \textsc{diamond} requires only 3 NFE per frame compared to 16 NFE per frame for \textsc{iris}. This is further reflected by the fact that \textsc{diamond} has significantly fewer parameters and takes less time to train than \textsc{iris}, as provided in Appendix \ref{app:performance_profile}.
\section{Scaling the diffusion world model to \textit{Counter-Strike: Global Offensive}\protect\footnote{This section was added after NeurIPS acceptance, following community interest in later CS:GO experiments.}}
\label{sec:csgo}

To investigate the ability of \textsc{diamond}'s diffusion world model to learn to model more complex 3D environments, we train the world model in isolation on static data from the popular video game \textit{Counter-Strike: Global Offensive} (CS:GO). We use the \textit{Online} dataset of 5.5M frames (95 hours) of online human gameplay captured at 16Hz from the map \textit{Dust II} by \citet{pearce2022counter}. We randomly hold out 0.5M frames (corresponding to 500 episodes, or 8 hours) for testing, and use the remaining 5M frames (87 hours) for training. There is no reinforcement learning agent or online data collection involved in these experiments.

To reduce the computational cost, we reduce the resolution from $(280\times150)$ to $(56\times30)$ for world modeling. We then introduce a second, smaller diffusion model as an upsampler to improve the generated images at the original resolution \citep{saharia2022image}. We scale the channels of the U-Net to increase the number of parameters from 4M for our Atari models to 381M for our CS:GO model (including 51M for the upsampler). The combined model was trained for 12 days on an RTX 4090. 

Finally, we introduce stochastic sampling and increase the number of denoising steps for the upsampler to 10, which we found to improve the resulting visual quality of the generations, while keeping the dynamics model the same (in particular, still using only 3 denoising steps). This enables a reasonable tradeoff between visual quality and inference cost, with the model running at 10Hz on an RTX 3090. Typical generations of the model are provided in Figure \ref{fig:csgo_grid} below.

%%%%%%%%%%%%%%%%%%%%%%%
\begin{figure}[h]
%\vspace{-6mm}
\begin{center}
\includegraphics[width=.7\linewidth]{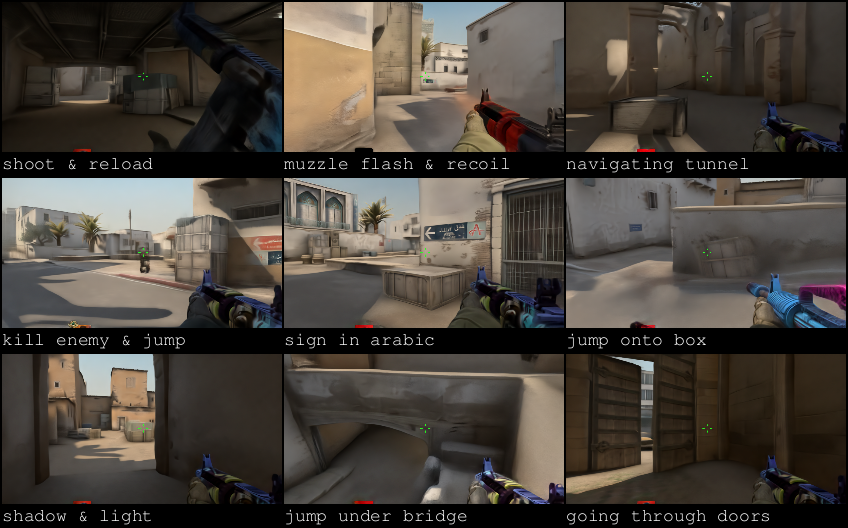}
\caption{Images captured from people playing with keyboard and mouse inside \textsc{diamond}'s diffusion world model. This model was trained on $87$ hours of static \textit{Counter-Strike: Global Offensive} (CS:GO) gameplay \citep{pearce2022counter} to produce an interactive neural game engine for the popular in-game map, \textit{Dust II}. Best viewed as videos at \wslink.}
\label{fig:csgo_grid}
\end{center}
\vspace{-4mm}
\end{figure}
%%%%%%%%%%%%%%%%%%%%%%%

We find the model is able to generate stable trajectories over hundreds of timesteps, although is more likely to drift out-of-distribution in less frequently visited areas of the map. Due to the limited memory of the model, approaching walls or losing visibility may cause the model to forget the current state and instead generate a new weapon or area of map. Interestingly, we find the model wrongly enables successive jumps by generalizing the effect of a jump on the geometry of the scene, since multiple jumps do not appear often enough in the training gameplay for the model to learn that mid-air jumps should be ignored. We expect scaling the model and data to address many of these limitations, with the exception of the memory of the model. Quantitative measurements of the capabilities of the CS:GO world model and attempts to address these limitations are left to future work.

\section{Related work}
\label{sec:related_work}

\textbf{World models.} The idea of reinforcement learning (RL) in the imagination of a neural network world model was introduced by \citet{ha2018world}. SimPLe \citep{kaiser2019atari100k} applied world models to Atari, and introduced the Atari 100k benchmark to focus on sample efficiency. Dreamer \citep{hafner2020dream} introduced RL from the latent space of a recurrent state space model (RSSM). DreamerV2 \citep{hafner2021mastering} demonstrated that using discrete latents could help to reduce compounding error, and DreamerV3 \citep{hafner2023dreamerv3} was able to achieve human-level performance on a wide range of domains with fixed hyperparameters. TWM \citep{robine2023transformer} adapts DreamerV2's RSSM to use a transformer architecture, while STORM \citep{zhang2023storm} adapts DreamerV3 in a similar way but with a different tokenization approach. Alternatively, IRIS \citep{iris2023} builds a language of image tokens with a discrete autoencoder, and composes these tokens over time with an autoregressive transformer. 

\textbf{Generative vision models.} There are parallels between these world models and image generation models which suggests that developments in generative vision models could provide benefits to world modeling. Following the rise of transformers in natural language processing \citep{vaswani2017attention,devlin2018bert,radford2019language}, VQGAN \citep{esser2021taming} and DALL·E \citep{ramesh2021zero} convert images to discrete tokens with discrete autoencoders \citep{vqvae}, and leverage the sequence modeling abilities of autoregressive transformers to build powerful text-to-image generative models. Concurrently, diffusion models \citep{sohl2015difforigin,ho2020DDPM,song_sde} gained traction \citep{dhariwal2021diffbeatsgans,ldm_stable_diffusion}, and have become a dominant paradigm for high-resolution image generation \citep{saharia2022imagen,ramesh2022hierarchical,podell2023sdxl}.

The same trends have taken place in the recent developments of video generation methods. VideoGPT \citep{yan2021videogpt} provides a minimal video generation architecture by combining a discrete autoencoder with an autoregressive transformer. Godiva \citep{wu2021godiva} enables text conditioning with promising generalization. Phenaki \citep{phenaki} allows arbitrary length video generation with sequential prompt conditioning. TECO \citep{yan2023teco} improves upon autoregressive modeling by using MaskGit \citep{chang2022maskgit}, and enables longer temporal dependencies by compressing input sequence embeddings. Diffusion models have also seen a resurgence in video generation using 3D U-Nets to provide high quality but short-duration video \citep{singer2023make,bar2024lumiere}. 
Recently, transformer-based diffusion models such as DiT  \citep{dit2023} and Sora \citep{sora2024} have shown improved scalability for both image and video generation, respectively.

\textbf{Diffusion for reinforcement learning.} There has also been much interest in combining diffusion models with reinforcement learning. This includes taking advantage of the flexibility of diffusion models as a policy \citep{wang2022diffusion, ajay2022conditional,pearce2023imitating}, as planners \citep{janner2022planning, liang2023adaptdiffuser}, as reward models \citep{nuti2023extracting}, and trajectory modeling for data augmentation in offline RL \citep{lu2023synthetic,ding2024diffusion,jackson2024policyguided}. \textsc{diamond} represents the first use of diffusion models as world models for learning online in imagination. 

\textbf{Generative game engines.} Playable games running entirely on neural networks have recently been growing in scope. \textit{GameGAN} \citep{gameGAN2020} learns generative models of games using a GAN \citep{goodfellow2014GAN} while \citet{bamford2020neural} use a Neural GPU \citep{kaiser2015neural}. Concurrent work includes \textit{Genie} \citep{bruce2024genie}, which generates playable platformer environments from image prompts, and \textit{GameNGen} \citep{valevski2024diffusionmodelsrealtimegame}, which similarly leverages a diffusion model to obtain a high resolution simulator of the game DOOM, but at a larger scale.
\vspace{-2mm}
\section{Limitations}
\label{sec:limitations}

We identify three main limitations of our work for future research. First, our main evaluation is focused on discrete control environments, and applying \textsc{diamond} to the continuous domain may provide additional insights. Second, the use of frame stacking for conditioning is a minimal mechanism to provide a memory of past observations. Integrating an autoregressive transformer over environment time, using an approach such as \citet{dit2023}, would enable longer-term memory and better scalability. We include an initial investigation into a potential cross-attention architecture in Appendix \ref{app:additonal_experiments}, but found frame-stacking more effective in our early experiments. Third, we leave potential integration of the reward/termination prediction into the diffusion model for future work, since combining these objectives and extracting representations from a diffusion model is not trivial \citep{luo2023dhf, xu2023open} and would make our world model unnecessarily complex.

\vspace{-2mm}
\section{Conclusion and Broader Impact}
\label{sec:conclusion}

We have introduced \textsc{diamond}, a reinforcement learning agent trained in a diffusion world model. 
We explained the key design choices we made to adapt diffusion for world modeling and to make our world model stable over long time horizons with a low number of denoising steps.
\textsc{diamond} achieves a mean human normalized score of $1.46$ on the well-established Atari 100k benchmark; a new best among agents trained entirely within a world model. 
We analyzed our improved performance in some games and found that it likely follows from better modeling of critical visual details.
We further demonstrated \textsc{diamond}'s diffusion world model can successfully model 3D environments and serve as a real-time neural game engine by training on static \textit{Counter-Strike: Global Offensive} gameplay.

World models constitute a promising direction to address sample efficiency and safety concerns associated with training agents in the real world. However, imperfections in the world model may lead to suboptimal or unexpected agent behaviors. We hope that the development of more faithful and interactive world models will contribute to broader efforts to further reduce these risks.

\newpage
 %\section*{Broader Impact}
% This paper considers the training of autonomous agents using a world model. The deployment of autonomous agents in the real world raises safety concerns regarding the potential harm that may be caused by an agent's actions. Training in simulation reduces these risks by reducing the time the agent spends interacting with the environment. However, imperfect world models may lead to unexpected behaviors in the real world. The development of more realistic world models should therefore reduce the risk associated with deploying agents trained in this manner. Additionally, as with all advances in the field of Machine Learning, there are many other potential societal consequences of our work, but none of which we feel must be specifically highlighted here.

\begin{ack}

We would like to thank Andrew Foong, Bálint Máté, Clément Vignac, Maxim Peter, Pedro Sanchez, Rich Turner, Stéphane Nguyen, Tom Lee, Trevor McInroe and Weipu Zhang for insightful discussions and comments.
Adam and Eloi met during an internship at Microsoft Research Cambridge, and would like to thank the Game Intelligence team, including Anssi Kanervisto, Dave Bignell, Gunshi Gupta, Katja Hofmann, Lukas Schäfer, Raluca Georgescu, Sam Devlin, Sergio Valcarcel Macua, Shanzheng Tan, Tabish Rashid, Tarun Gupta, Tim Pearce, and Yuhan Cao, for their support in the early stages of this project, and a great summer.

\end{ack}

\bibliography{main}
\bibliographystyle{apalike}

%%%%%%%%%%%%%%%%%%%%%%%%%%%%%%%%%%%%%%%%%%%%%%%%%%%%%%%%%%%%

\newpage
\appendix
\section{Sampling observations in \textsc{diamond}}
\label{appendix:sampling}

We describe here how we sample an observation $\x_t^0$ from our diffusion world model. We initialize the procedure with a noisy observation $\x_t^\Tau \sim p^{prior}$, and iteratively solve the reverse SDE in Equation \ref{eq:reverse_process} from $\tau = \Tau$ to $\tau = 0$, using the learned score model $\mathbf{S}_\theta(\x_t^\tau, \tau, \x_{<t}^0, a_{<t})$ conditioned on past observations $\x_{<t}^0$ and actions $a_{<t}$. This procedure is illustrated in Figure \ref{fig:architecture}.

In fact, there are many possible sampling methods for a given learned score model $\mathbf{S}_\theta$ \citep{karras2022elucidating}. Notably, \citet{song_sde} introduce a corresponding ``probability flow" ordinary differential equation (ODE), with marginals equivalent to the stochastic process described in Section \ref{subsec:diffusion}. In that case, the solving procedure is deterministic, and the only randomness comes from sampling the initial condition. In practice, this means that for a given score model, we can resort to any ODE or SDE solver, from simple first order methods like Euler (deterministic) and Euler–Maruyama (stochastic) schemes, to higher-order methods like Heun's method \citep{ascher1998computer}. 

Regardless of the choice of solver, each step introduces truncation errors, resulting from the local score approximation and the discretization of the continuous process. Higher order samplers may reduce this truncation error, but come at the cost of additional Number of Function Evaluations (NFE) -- how many forward passes of the network are required to generate a sample. This local error generally scales superlinearly with respect to the step size (for instance Euler's method is $\mathcal{O}(h^2)$ for step size $h$), so increasing the number of denoising steps improves the visual quality of the generated next frame. Therefore, there is a trade-off between visual quality and NFE that directly determines the inference cost of the diffusion world model.

\section{Link between DDPM and continuous-time score-based diffusion models}
\label{app:ddpm}

Denoising Diffusion Probabilistic Models (\textsc{ddpm}, \citet{ho2020DDPM}) can be described as a discrete version of the diffusion process introduced in Section \ref{subsec:diffusion}, as described in \citet{song_sde}. The discrete forward process is a Markov chain characterized by a discrete noise schedule $0 < \beta_1, \dots, \beta_i, \dots \beta_N < 1$, and a variance-preserving Gaussian transition kernel,

\begin{equation}
    p(\x^i|\x^{i-1}) = \mathcal{N}(\x^i; \sqrt{1-\beta_i} \x^{i-1}, \beta_i \mathbf{I}).
\end{equation}

In the continuous time limit $N \to \infty$, the Markov chain becomes a diffusion process, and the discrete noise schedule becomes a time-dependent function $\beta(\tau)$. This diffusion process can be described by an SDE with drift coefficient $\mathbf{f}(\x, \tau) = -\frac{1}{2}\beta(\tau)\x$ and diffusion coefficient $g(\tau) = \sqrt{\beta(\tau)}$ \citep{song_sde}.

\section{\textsc{EDM} network preconditioners and training}
\label{appendix:karras_conditioners}

\citet{karras2022elucidating} use the following preconditioners for normalization and rescaling purposes (as mentioned in Section \ref{subsec:practical_dwm}) to improve network training:

\begin{equation}
    c_{in}^\tau = \frac{1}{\sqrt{\sigma(\tau)^2 + \sigma_{data}^2}}
\end{equation}
\begin{equation}
    c_{out}^\tau = \frac{\sigma(\tau)\sigma_{data}}{\sqrt{\sigma(\tau)^2 + \sigma_{data}^2}}
\end{equation}
\begin{equation}
    c_{noise}^\tau = \frac{1}{4}\log(\sigma(\tau))
\end{equation}
\begin{equation}
    c_{skip}^\tau = \frac{\sigma_{data}^2}{\sigma_{data}^2 + \sigma^2(\tau)},
\end{equation}
where $\sigma_{data}=0.5$.

The noise parameter $\sigma(\tau)$ is sampled to maximize the effectiveness of training as follows:
\begin{equation}
\log(\sigma(\tau))\sim \mathcal{N}(P_{mean}, P_{std}^2),
\end{equation}
where $P_{mean}=-0.4, P_{std}=1.2$. Refer to \citet{karras2022elucidating} for an in-depth analysis.
\newpage
\section{Model architectures}\label{app:architectures}

The diffusion model $\mathbf{D}_\theta$ is a standard U-Net 2D \citep{ronneberger2015unet}, conditioned on the last 4 frames and actions, as well as the diffusion time $\tau$. We use frame stacking for observation conditioning, and adaptive group normalization \citep{adagn} for action and diffusion time conditioning.

The reward/termination model $R_\psi$ layers are shared except for the final prediction heads. The model takes as input a sequence of frames and actions, and forwards it through convolutional residual blocks \citep{He2015} followed by an LSTM cell \citep{a3c,lstm,Gers2000}. Before starting the imagination procedure, we burn-in \citep{r2d2} the conditioning frames and actions to initialize the hidden and cell states of the LSTM. 

The weights of the policy $\pi_\phi$ and value network $V_\phi$ are shared except for the last layer. In the following, we refer to $(\pi,V)_\phi$ as the "actor-critic" network, even though $V$ is technically a state-value network, not a critic. This network takes as input a frame, and forwards it through convolutional trunk followed by an LSTM cell. The convolutional trunk consists of four residual blocks and 2x2 max-pooling with stride 2. The main path of the residual blocks consists of a group normalization \citep{groupnorm} layer, a SiLU activation \citep{elfwing2018sigmoid}, and a 3x3 convolution with stride 1 and padding 1. Before starting the imagination procedure, we burn-in the conditioning frames to initialize the hidden and cell states of the LSTM.

Please refer to Table \ref{tbl_architecture} below for hyperparameter values, and to Algorithm \ref{alg:diamond} for a detailed summary of the training procedure. 

\vspace{1cm}

\begin{table}[h!]
\caption{Architecture details for \textsc{diamond}.}
\label{tbl_architecture}
\begin{center}
% \resizebox{0.4 \columnwidth}{!}{
\begin{tabular}{ l c }
\multicolumn{1}{c}{\textbf{Hyperparameter}}  & \multicolumn{1}{c}{\textbf{Value}} \\ 

\hline \\

% \hline \\
\\
\multicolumn{2}{l}{\textbf{Diffusion Model ($\mathbf{D}_\theta$)}} \\
% Number of conditioning observations/actions ($L$) & 4 \\
Observation conditioning mechanism & Frame stacking \\
Action conditioning mechanism & Adaptive Group Normalization \\
Diffusion time conditioning mechanism & Adaptive Group Normalization \\
Residual blocks layers & [2, 2, 2, 2] \\
Residual blocks channels & [64, 64, 64, 64] \\
Residual blocks conditioning dimension & 256 \\

% \hline \\
\\
\multicolumn{2}{l}{\textbf{Reward/Termination Model ($R_\psi$)}} \\
Action conditioning mechanisms & Adaptive Group Normalization \\
Residual blocks layers & [2, 2, 2, 2] \\
Residual blocks channels & [32, 32, 32, 32] \\
Residual blocks conditioning dimension & 128 \\
LSTM dimension & 512 \\ 
% Burn-in length ($B_{rt}$), set as $B_{rt} = L$ in practice & 4 \\ 
% Training sequence length ($B_{rt} + H$) & 19 \\

% \hline \\
\\
\multicolumn{2}{l}{\textbf{Actor-Critic Model ($\pi_\phi$ and $V_\phi$)}} \\
Residual blocks layers & [1, 1, 1, 1] \\
Residual blocks channels & [32, 32, 64, 64] \\
LSTM dimension & 512 \\ 
% Burn-in length ($B_{ac}$), set as $B_{ac} = L$ in practice & 4 \\

\end{tabular}
% }
\end{center}
\end{table}

\newpage
\section{Training hyperparameters}
\label{app:hyperparams}

\vspace{1cm}

\begin{table}[h!]
\caption{Hyperparameters for \textsc{diamond}.}
\label{tbl_atari_hypers}
\begin{center}
% \resizebox{0.4 \columnwidth}{!}{
\begin{tabular}{ l c }
\multicolumn{1}{c}{\textbf{Hyperparameter}}  & \multicolumn{1}{c}{\textbf{Value}} \\ 

\hline \\

\multicolumn{2}{l}{\textbf{Training loop}} \\
Number of epochs & 1000 \\
Training steps per epoch & 400 \\
Batch size & 32 \\
Environment steps per epoch & 100 \\
Epsilon (greedy) for collection & 0.01 \\

\\

\multicolumn{2}{l}{\textbf{RL hyperparameters}} \\
Imagination horizon ($H$) & 15 \\
Discount factor ($\gamma$) &  0.985 \\ 
Entropy weight ($\eta$) & 0.001 \\ 
$\lambda$-returns coefficient ($\lambda$) & 0.95 \\

\\

\multicolumn{2}{l}{\textbf{Sequence construction during training}} \\
For $\mathbf{D}_\theta$, number of conditioning observations and actions ($L$) & 4 \\
For $R_\psi$, burn-in length ($B_{R}$), set to $L$ in practice & 4 \\ 
For $R_\psi$, training sequence length ($B_{R} + H$) & 19 \\
For $\pi_\phi$ and $V_\phi$, burn-in length ($B_{\pi,V}$), set to $L$ in practice & 4 \\

\\
\multicolumn{2}{l}{\textbf{Optimization}} \\
Optimizer & AdamW \\
Learning rate & 1e-4 \\
Epsilon & 1e-8 \\
Weight decay ($\mathbf{D}_\theta$) & 1e-2 \\
Weight decay ($R_\psi$) & 1e-2 \\
Weight decay ($\pi_\phi$ and $V_\phi$)  & 0 \\

% \hline \\
\\
\multicolumn{2}{l}{\textbf{Diffusion Sampling}} \\
Method &  Euler \\
Number of steps & 3 \\

\\
\multicolumn{2}{l}{\textbf{Environment}} \\
Image observation dimensions & 64$\times$64$\times$3 \\
Action space & Discrete (up to 18 actions) \\
Frameskip & 4 \\
Max noop & 30 \\
Termination on life loss & True \\
Reward clipping & $\{-1, 0, 1\}$ \\

\end{tabular}
% }
\end{center}
\end{table}

\newpage
\section{Reinforcement learning objectives}
\label{appendix:rl_actor_critic}

In what follows, we note $\x_t$, $r_t$ and $d_t$ the observations, rewards, and boolean episode terminations predicted by our world model. We note $H$ the imagination horizon, $V_\phi$ the value network, $\pi_\phi$ the policy network, and $a_t$ the actions taken by the policy within the world model. 

We use $\lambda$-returns to balance bias and variance as the regression target for the value network. Given an imagined trajectory of length $H$, we can define the $\lambda$-return recursively as follows,

\begin{equation}
\Lambda_t = 
\begin{cases}
    r_t + \gamma (1 - d_t) \Big[ (1 - \lambda) V_\phi(\x_{t+1}) + \lambda \Lambda_{t+1} \Big]   & \text{if}\quad t < H \\
    V_\phi(\x_H)                                                                                            & \text{if}\quad t = H. \\
\end{cases}
\end{equation}

The value network $V_\phi$ is trained to minimize $\mathcal{L}_V(\phi)$, the expected squared difference with $\lambda$-returns over imagined trajectories,

\begin{equation}
\mathcal{L}_V(\phi) = \mathbb{E}_{\pi_\phi} \left[ \sum_{t=0}^{H-1} \big( V_\phi(\x_t) - \mathrm{sg} ( \Lambda_t ) \big)^2 \right],
\end{equation}

where $\operatorname{sg}(\cdot)$ denotes the gradient stopping operation, meaning that the target is a constant in the gradient-based optimization, as classically established in the literature \citep{mnih2015dqn,hafner2021mastering,iris2023}.

As we can generate large amounts of on-policy trajectories in imagination, we use a simple \textsc{reinforce} objective to train the policy, with the value $V_\phi(\x_t)$ as a baseline to reduce the variance of the gradients \citep{sutton2018reinforcement}. The policy is trained to minimize the following objective, combining \textsc{reinforce} and a weighted entropy maximization objective to maintain sufficient exploration,

\begin{equation}
\mathcal{L}_\pi(\phi) = - \mathbb{E}_{\pi_\phi} \left[ \sum_{t=0}^{H-1} \log\left(\pi_\phi\left(a_t \mid \x_{\le t}\right)\right) \operatorname{sg}\left(\Lambda_t - V_\phi\left(\x_t\right)\right) + \eta \operatorname{\mathcal{H}}\left(\pi_\phi \left(a_t \mid \x_{\le t} \right) \right)\right].
\end{equation}

\vspace{1cm}

\newpage
\section{\textsc{diamond} algorithm}
\label{app:algorithm}

We summarize the overall training procedure of \textsc{diamond} in Algorithm \ref{alg:diamond} below. We denote as $\mathcal{D}$ the replay dataset where the agent stores data collected from the real environment, and other notations are introduced in previous sections or are self-explanatory.

\begin{algorithm}[H]
\caption{\textsc{diamond}}
\label{alg:diamond}
\DontPrintSemicolon

\SetKwProg{Proc}{Procedure}{:}{}
\SetKwFunction{FTrain}{training\_loop}
\SetKwFunction{FCollect}{collect\_experience}
\SetKwFunction{FUpdateDiffusionModel}{update\_diffusion\_model}
\SetKwFunction{FUpdateRewardEndModel}{update\_reward\_end\_model}
\SetKwFunction{FUpdateActorCritic}{update\_actor\_critic}

\Proc{\FTrain{}}
{
    \For{epochs}{
        \texttt{collect\_experience(}\textit{steps\_collect}\texttt{)} \;
        \For{steps\_diffusion\_model}{
            \texttt{update\_diffusion\_model()} \; 
        }
        \For{steps\_reward\_end\_model}{
            \texttt{update\_reward\_end\_model()} \; 
        }
        \For{steps\_actor\_critic}{
            \texttt{update\_actor\_critic()} \; 
        }
    }
}

\Proc{\FCollect{$n$}}
{
    $\x_0^0 \gets \texttt{env.reset()}$ \;
    \For{$t = 0$ \KwTo $n - 1$}{
        Sample $a_t \sim \pi_\phi(a_t \mid \x_t^0)$ \;
        $\x_{t+1}^0, r_t, d_t \gets \texttt{env.step(}a_t\texttt{)}$ \;
        $\mathcal{D} \gets \mathcal{D} \cup \{ \x_t^0, a_t, r_t, d_t \} $ \;
        \If{$d_t = 1$}{
            $\x_{t+1}^0 \gets \texttt{env.reset()}$ \;
        }
    }
}

\Proc{\FUpdateDiffusionModel{}}
{
    Sample sequence $ ( \x_{t-L+1}^0, a_{t-L+1}, \dots, \x_t^0, a_t, \x_{t+1}^0 ) \sim \mathcal{D} $ \;
    Sample $\log(\sigma) \sim \mathcal{N}(P_{mean}, P_{std}^2)$ \tcp*[f]{log-normal sigma distribution from EDM} \;
    Define $\tau := \sigma$\tcp*[f]{default identity schedule from EDM} \;
    Sample $\x_{t+1}^\tau \sim \mathcal{N}(\x_{t+1}^0, \sigma^2 \mathbf{I})$ \tcp*[f]{Add independent Gaussian noise} \;
    Compute $\hat{\x}_{t+1}^0 = \mathbf{D}_\theta(\x_{t+1}^\tau, \tau, \x_{t-L+1}^0, a_{t-L+1}, \dots, \x_t^0, a_t)$ \;
    Compute reconstruction loss $\mathcal{L}(\theta) = \Vert \hat{\x}_{t+1}^0 - \x_{t+1}^0 \Vert^2$ \;
    Update $\mathbf{D}_\theta$ \;
}

\Proc{\FUpdateRewardEndModel{}}
{
    Sample indexes $\mathcal{I} \coloneqq \{t, \dots, t+L+H-1 \}$ \tcp*[f]{burn-in + imagination horizon} \;
    Sample sequence $ ( \x_i^0, a_i, r_i, d_i )_{i \in \mathcal{I}} \sim \mathcal{D} $ \;
    Initialize $h = c = 0$ \tcp*[f]{LSTM hidden and cell states}\;  
    \For{$i \in \mathcal{I}$}{
        Compute $\hat{r}_i, \hat{d}_i, h, c = R_\psi(\x_i, a_i, h, c)$ \;
    }
    Compute $\mathcal{L}(\psi) = \sum_{i \in \mathcal{I}} \mathrm{CE}(\hat{r}_i, \mathrm{sign}(r_i)) + \mathrm{CE}(\hat{d}_i, d_i)$ \tcp*[f]{CE: cross-entropy loss} \;
    Update $R_\psi$ \;
}

\Proc{\FUpdateActorCritic{}}
{
    Sample initial buffer $( \x_{t-L+1}^0, a_{t-L+1}, \dots, \x_t^0) \sim \mathcal{D} $ \;
    Burn-in buffer with $R_\psi$, $\pi_\phi$ and $V_\phi$ to initialize LSTM states \;
    \For{$i=t$ \KwTo $t + H - 1$}{
        Sample  $a_i \sim \pi_\phi(a_i \mid \x_i^0)$ \;
        Sample reward $r_i$ and termination $d_i$ with $ R_\psi$ \;
        Sample next observation $\x_{i+1}^0$ by simulating reverse diffusion process with $\mathbf{D}_\theta$ \;
    }
    Compute $ V_\phi(\x_i) $ for $i = t, \dots, t + H$ \;
    Compute RL losses $\mathcal{L}_V(\phi)$ and $\mathcal{L}_\pi(\phi)$ \;
    Update $\pi_\phi$ and $V_\phi$ \;
}
\end{algorithm}
\newpage
\section{Additional performance comparisons}
\label{app:performance_profile}

We provide performance profiles \citep{agarwal2021deep} for \textsc{diamond} and baselines below. % We see that \textsc{diamond} outperforms the baselines in terms of the fraction of runs above a given human normalized score for a range of scores.

%%%%%%%%%%%%%%%%%%%%%%%
\begin{figure}[h!]
\vskip 0.2in
\begin{center}
\centerline{\includegraphics[width=.7\columnwidth]{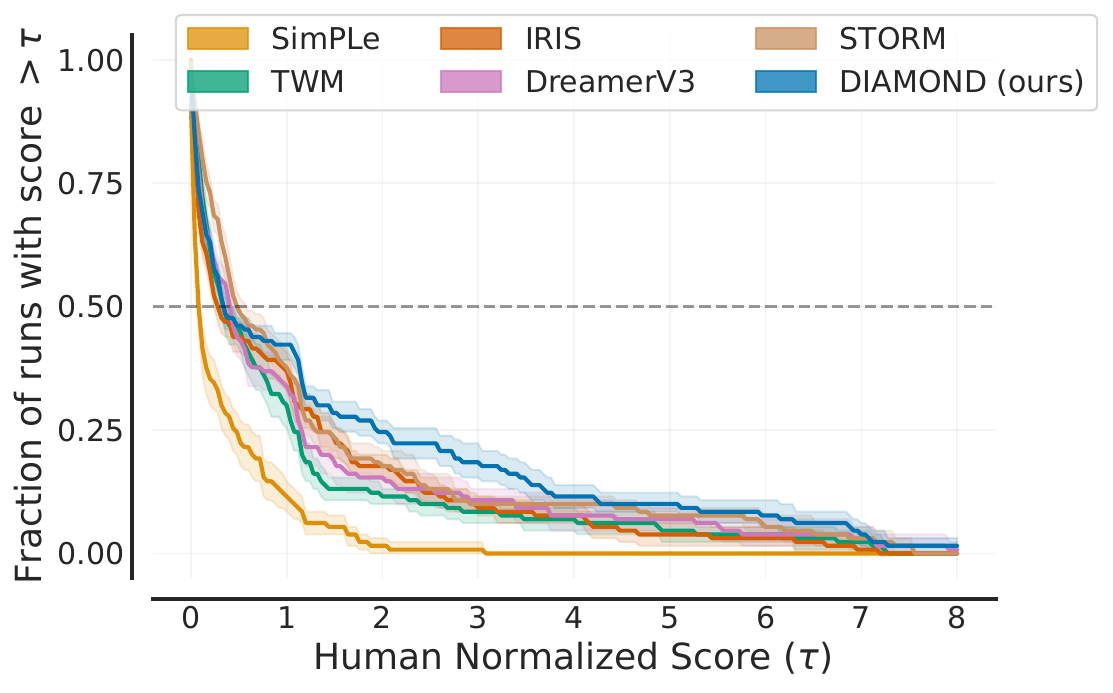}}
\caption{Performance profiles, i.e. fraction of runs above a given human normalized score.}
\label{fig:results_performance_profile}
\end{center}
\vskip 0.2in
\end{figure}
%%%%%%%%%%%%%%%%%%%%%%%

As additional angles of comparison, we also provide parameter counts and approximate training times for \textsc{iris}, DreamerV3 and \textsc{diamond} in Table \ref{tab:training_time} below. We see that \textsc{diamond} has the highest mean HNS, with fewer parameters than both \textsc{iris} and DreamerV3. \textsc{diamond} also trains faster than \textsc{iris}, although is slower than DreamerV3. 

\begin{table}[h]
\caption{Number of parameters, training time, and mean human-normalized score (HNS).}
\label{tab:training_time}
\begin{center}
\begin{tabular}{lcccc}
\hline
    & \textsc{iris} & DreamerV3 & \textsc{diamond} (ours) & \\ \hline
\#parameters (↓)        & 30M       & 18M            & \textbf{13M}   &  \\
Training days (↓) & 4.1       & \textbf{\textless 1}   & 2.9   &  \\
Mean HNS (↑)            & 1.046     & 1.097          & \textbf{1.459} &  \\ \hline
\end{tabular}
\end{center}
\end{table}

A full training time profile for \textsc{diamond} is provided in Appendix~\ref{app:profiling}.
\newpage
\section{Training time profile}\label{app:profiling}

Table~\ref{tab:profiling} provides a full training time profile for \textsc{diamond}.

\begin{table}[h!]
\caption{Detailed breakdown of training time. Profiling performed using a Nvidia RTX 4090 with the default hyperparameters specified in Appendices \ref{app:architectures} and \ref{app:hyperparams} These profiling measures are representative, since exact durations will depend on the machine, the environment, and the training stage.}
\label{tab:profiling}
\begin{center}

\scalebox{1.0}{

\begin{tabular}{|>{\raggedright\arraybackslash}p{7cm}|>{\raggedleft\arraybackslash}p{2cm}|>{\raggedleft\arraybackslash}p{3.5cm}|}
    \hline
    \textbf{Single update} & \textbf{Time (ms)} & \textbf{Detail (ms)} \\
    \hline
    Total & $543$ & $88 + 115 + 340$\\
    \hspace{0.5cm} Diffusion model update & $88$ & - \\
    \hspace{0.5cm} Reward/Termination model update & $115$ & - \\
    \hspace{0.5cm} Actor-Critic model update & $340$ & $15 \times 20.4 + 34$\\
    \hspace{1cm} Imagination step (x 15) & $20.4$ & $12.7 + 7.0 + 0.7$ \\
    \hspace{1.5cm} Next observation prediction & $12.7$ & $3 \times 4.2$ \\
    \hspace{2cm} Denoising step (x 3) & $4.2$ & - \\
    \hspace{1.5cm} Reward/Termination prediction & $7.0$ & - \\
    \hspace{1.5cm} Action prediction & $0.7$ & - \\
    \hspace{1cm} Loss computation and backward & $34$ & - \\
    \hline
    \textbf{Epoch} & \textbf{Time (s)} & \textbf{Detail (s)} \\
    \hline
    Total & $217$ & $35 + 46 + 136$ \\
    \hspace{0.5cm} Diffusion model & $35$ & $400 \times 88 \times  10^{-3}$ \\
    \hspace{0.5cm} Reward/Termination model & $46$ & $400 \times 115 \times  10^{-3}$ \\
    \hspace{0.5cm} Actor-Critic model & $136$ & $400 \times 340 \times  10^{-3}$ \\
    \hline
    \textbf{Run} & \textbf{Time (days)} & \textbf{Detail (days)} \\
    \hline
    Total & $2.9$ & $2.5 + 0.4$ \\
    \hspace{0.5cm} Training time & $2.5$ & $1000 \times 217 / (24 \times 3600)$ \\
    \hspace{0.5cm} Other (collection, evaluation, checkpointing) & $0.4$ & - \\
    \hline
\end{tabular}

}

\end{center}
\end{table}
\label{tab:r}
\newpage
\section{Broader comparison to model-free and search-based methods}
\label{app:additonal_baselines}

Table \ref{tab:atari_results_other_baselines} provides scores for model-free and search-based methods, including the current best performing methods on the Atari 100k benchmark, EfficientZero \citep{ye2021efficientzero} and \textsc{bbf} \citep{schwarzer2023bigger}. Both of these methods use approaches that are out of scope of our approach, such as computationally expensive lookahead Monte-Carlo tree search for EfficientZero, and using periodic network resets in combination with hyperparameter scheduling for \textsc{bbf}. We see that while the use of lookahead search and more advanced reinforcement learning techniques (for EfficientZero \citep{ye2021efficientzero} and \textsc{bbf} \citep{schwarzer2023bigger} respectively) can still provide greater performance overall, \textsc{diamond} promisingly still outperforms these methods on some games.

\begin{table*}[h]
    \caption{Raw scores and human-normalized metrics for search-based and model-free methods.}
    \label{tab:atari_results_other_baselines}
\begin{center}
\begin{small}
\centering
\scalebox{0.86}{
\centering

\begin{tabular}{lrr rrrrrr}
\toprule
\multicolumn{2}{c}{} & \multicolumn{2}{c}{Search-based} & \multicolumn{4}{c}{Model-free} \\
\cmidrule(lr){3-4} \cmidrule(lr){5-8}
%%%%%%%%%%%%%%%%%%%%%%%%%%%%%%%%%%%%%%%%%%%%%%%%%%%%%%%%%%%%%%%%%%%%
Game                 &  Human     &  MuZero    &  EfficientZero      &  CURL     &  SPR       &  SR-SPR              &  BBF                &  \textsc{diamond} (ours)  \\
\midrule
Alien                &  7127.7    &  530.0     &  808.5              &  711.0    &  841.9     &  1107.8              &  \textbf{1173.2}    &  744.1                    \\
Amidar               &  1719.5    &  38.8      &  148.6              &  113.7    &  179.7     &  203.4               &  \textbf{244.6}     &  225.8                    \\
Assault              &  742.0     &  500.1     &  1263.1             &  500.9    &  565.6     &  1088.9              &  \textbf{2098.5}    &  1526.4                   \\
Asterix              &  8503.3    &  1734.0    &  \textbf{25557.8}   &  567.2    &  962.5     &  903.1               &  3946.1             &  3698.5                   \\
BankHeist            &  753.1     &  192.5     &  351.0              &  65.3     &  345.4     &  531.7               &  \textbf{732.9}     &  19.7                     \\
BattleZone           &  37187.5   &  7687.5    &  13871.2            &  8997.8   &  14834.1   &  17671.0             &  \textbf{24459.8}   &  4702.0                   \\
Boxing               &  12.1      &  15.1      &  52.7               &  0.9      &  35.7      &  45.8                &  85.8               &  \textbf{86.9}            \\
Breakout             &  30.5      &  48.0      &  \textbf{414.1}     &  2.6      &  19.6      &  25.5                &  370.6              &  132.5                    \\
ChopperCommand       &  7387.8    &  1350.0    &  1117.3             &  783.5    &  946.3     &  2362.1              &  \textbf{7549.3}    &  1369.8                   \\
CrazyClimber         &  35829.4   &  56937.0   &  83940.2            &  9154.4   &  36700.5   &  45544.1             &  58431.8            &  \textbf{99167.8}         \\
DemonAttack          &  1971.0    &  3527.0    &  13003.9            &  646.5    &  517.6     &  2814.4              &  \textbf{13341.4}   &  288.1                    \\
Freeway              &  29.6      &  21.8      &  21.8               &  28.3     &  19.3      &  25.4                &  25.5               &  \textbf{33.3}            \\
Frostbite            &  4334.7    &  255.0     &  296.3              &  1226.5   &  1170.7    &  \textbf{2584.8}     &  2384.8             &  274.1                    \\
Gopher               &  2412.5    &  1256.0    &  3260.3             &  400.9    &  660.6     &  712.4               &  1331.2             &  \textbf{5897.9}          \\
Hero                 &  30826.4   &  3095.0    &  \textbf{9315.9}    &  4987.7   &  5858.6    &  8524.0              &  7818.6             &  5621.8                   \\
Jamesbond            &  302.8     &  87.5      &  517.0              &  331.0    &  366.5     &  389.1               &  \textbf{1129.6}    &  427.4                    \\
Kangaroo             &  3035.0    &  62.5      &  724.1              &  740.2    &  3617.4    &  3631.7              &  \textbf{6614.7}    &  5382.2                   \\
Krull                &  2665.5    &  4890.8    &  5663.3             &  3049.2   &  3681.6    &  5911.8              &  8223.4             &  \textbf{8610.1}          \\
KungFuMaster         &  22736.3   &  18813.0   &  \textbf{30944.8}   &  8155.6   &  14783.2   &  18649.4             &  18991.7            &  18713.6                  \\
MsPacman             &  6951.6    &  1265.6    &  1281.2             &  1064.0   &  1318.4    &  1574.1              &  \textbf{2008.3}    &  1958.2                   \\
Pong                 &  14.6      &  -6.7      &  20.1               &  -18.5    &  -5.4      &  2.9                 &  16.7               &  \textbf{20.4}            \\
PrivateEye           &  69571.3   &  56.3      &  96.7               &  81.9     &  86.0      &  97.9                &  40.5               &  \textbf{114.3}           \\
Qbert                &  13455.0   &  3952.0    &  \textbf{13781.9}   &  727.0    &  866.3     &  4044.1              &  4447.1             &  4499.3                   \\
RoadRunner           &  7845.0    &  2500.0    &  17751.3            &  5006.1   &  12213.1   &  13463.4             &  \textbf{33426.8}   &  20673.2                  \\
Seaquest             &  42054.7   &  208.0     &  1100.2             &  315.2    &  558.1     &  819.0               &  \textbf{1232.5}    &  551.2                    \\
UpNDown              &  11693.2   &  2896.9    &  17264.2            &  2646.4   &  10859.2   &  \textbf{112450.3}   &  12101.7            &  3856.3                   \\
\midrule
\#Superhuman (↑)     &  N/A       &  5         &  \textbf{14}        &  2        &  6         &  9                   &  12                 &  11                       \\
Mean (↑)             &  1.000     &  0.562     &  1.943              &  0.261    &  0.616     &  1.271               &  \textbf{2.247}     &  1.459                    \\
IQM (↑)              &  1.000     &  0.288     &  1.047              &  0.113    &  0.337     &  0.700               &  \textbf{1.139}     &  0.641                    \\
%%%%%%%%%%%%%%%%%%%%%%%%%%%%%%%%%%%%%%%%%%%%%%%%%%%%%%%%%%%%%%%%%%

\bottomrule
\end{tabular}
 }
\end{small}
\end{center}
%\vspace{-0.02\linewidth}
\end{table*}

\newpage
\section{Quantitative analysis of autoregressive model drift}\label{app:ddpm_drift}

\begin{figure}[h!]
\begin{center}
\centerline{\includegraphics[width=\columnwidth]{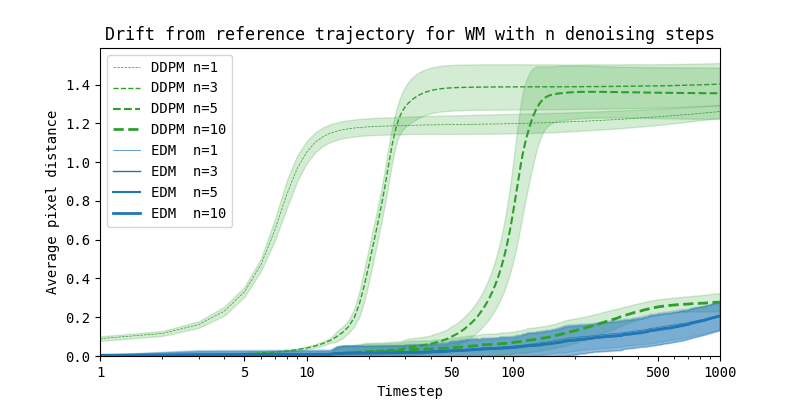}}
\caption{Average pixel drift between an imagined trajectory and the corresponding reference trajectory collected with an expert in \textit{Breakout}. The trajectories are each 1000 timesteps, starting from the same frame and following the same sequence of actions. Each line displays the average and shaded standard deviation of 400 reference trajectories held out from training data. DDPM becomes more stable with increasing number of denoising steps, but is less stable than 1-step EDM, even with 10 denoising steps. The drift we observe for EDM corresponds to differences in the imagined trajectory rather than a pathological color shift as we see in Figure 3a.}
\label{fig:ddpm_drift}
\end{center}
\end{figure}

Figure~\ref{fig:ddpm_drift} provides a quantitative measure of the compounding error demonstrated qualitatively in Figure~\ref{fig:denoising_trajectories} for DDPM and EDM based world models.

\section{Quantitative ablation on reducing the number of denoising steps}\label{app:denoising_ablation}

Table~\ref{tab:ablation_rl_1_step} provides a quantitative ablation of the effect of reducing the number of denoising steps used for our EDM diffusion world model from 3 (used for Table~\ref{tab:atari_results_full}) to 1, for \textsc{diamond}'s 10 highest performing games. Note that the 1-step results correspond to a single seed only so will have higher variance. Nonetheless, these results provide some signal that agents trained with 1 denoising step perform worse than our default choice of 3, particularly for the game \textit{Boxing}, despite the apparent similarity in Figure~\ref{fig:ddpm_drift}. This additional evidence supports our qualitative analysis in Section \ref{subsec:denoising_steps}.

\begin{table*}[h]
\caption{Quantitative ablation on reducing the number of denoising steps from 3 (default) to 1.}
\label{tab:ablation_rl_1_step}
\begin{center}
\begin{small}
\centering
\scalebox{1.0}{
\centering
\begin{tabular}{lrr rr}
\toprule
%\multicolumn{3}{c}{} & \multicolumn{2}{c}{Model-free} & \multicolumn{5}{c}{Imagination-based} \\
%\cmidrule(lr){4-5} \cmidrule(lr){6-11}

%%%%%%%%%%%%%%%%%%%%%%%%%%%%%%%%%%%%%%%%%%%%%%%%%%%%%%%%%%%%%%%%%%%%
Game                 &  Random    &  Human     &  \textsc{diamond} ($n=3$)  & \textsc{diamond} ($n=1$)  \\
\midrule
Amidar               &  5.8       &  1719.5    &  \textbf{225.8}           & 191.8 \\
Assault              &  222.4     &  742.0     &  \textbf{1526.4}          & 782.5 \\
Asterix              &  210.0     &  8503.3    &          3698.5           & \textbf{6687.0} \\
Boxing               &  0.1       &  12.1      &  \textbf{86.9}            & 41.9            \\
Breakout             &  1.7       &  30.5      &  \textbf{132.5}           & 50.8            \\
CrazyClimber         &  10780.5   &  35829.4   &  \textbf{99167.8}         & 87233.0         \\
Kangaroo             &  52.0      &  3035.0    &  \textbf{5382.2}          & 1710.0          \\
Krull                &  1598.0    &  2665.5    &          8610.1           & \textbf{9105.1} \\
Pong                 &  -20.7     &  14.6      &          20.4             & \textbf{20.9}   \\
RoadRunner           &  11.5      &  7845.0    &  \textbf{20673.2}         & 5084.0          \\
\midrule
Mean HNS (↑)         &  0.000     &  1.000     &  \textbf{3.052}           & 1.962           \\
%%%%%%%%%%%%%%%%%%%%%%%%%%%%%%%%%%%%%%%%%%%%%%%%%%%%%%%%%%%%%%%%%%

\bottomrule
\end{tabular}
 }
\end{small}
\end{center}
%\vspace{-1cm}
\end{table*}
\newpage
\newpage
\section{Early investigations on visual quality in more complex environments}
\label{app:additonal_experiments}

In the main body of the paper, we evaluated the utility of \textsc{diamond} for the purpose of training RL agents in a world model on the well-established Atari 100k benchmark \citep{kaiser2019atari100k}, and demonstrated \textsc{diamond}'s diffusion world model could be applied to model a more complex 3D environment from the game \textit{Counter-Strike: Global Offensive}. In this section, we provide early experiments investigating the effectiveness of \textsc{diamond}'s diffusion world model by directly evaluating the visual quality of the trajectories they generate. The two environments we consider are presented in Section \ref{app:subsec:environments} below.

\subsection{Environments}
\label{app:subsec:environments}

\textbf{CS:GO.} 
We use the \textit{Counter-Strike: Global Offensive} dataset introduced by \citet{pearce2022counter}. Here we use the \textit{Clean} dataset containing 190k frames (3.3 hours) of high-skill human gameplay, captured on the \textit{Dust II} map. This contains observations and actions (mouse and keyboard) captured at 16Hz. We use 150k frames (2.6 hours) for training and 40k frames (0.7 hours) for evaluation. We resize observations to 64$\times$64 pixels, and use no augmentation.
%\footnote{\href{https://github.com/TeaPearce/Counter-Strike_Behavioural_Cloning}{\texttt{https://github.com/TeaPearce/Counter\-Strike\_Behavioural\_Cloning}}}.

\textbf{Motorway driving.} We use the dataset from \citet{santana2016learning}\footnote{\href{https://github.com/commaai/research}{\texttt{https://github.com/commaai/research}}}, which contains camera and metadata captured from human drivers on US motorways. We select only trajectories captured in daylight, and exclude the first and last 5 minutes of each trajectory (typically traveling to/from a motorway), leaving 4.4 hours of data. We use five trajectories for training (3.6 hours) and two for testing (0.8 hours). We downsample the dataset to 10Hz, resize observations to 64$\times$64, and for actions use the (normalized) steering angle and acceleration. During training, we apply data augmentation of shift \& scale, contrast, brightness, and saturation, and mirroring.

We note that the purpose of our investigation is to train and evaluate \textsc{diamond}'s diffusion model on these static datasets, and that we do not perform reinforcement learning, since there is no standard reinforcement learning protocol for these environments.

\subsection{Diffusion Model Architectures}

\begin{figure}[h]
    \begin{center}
    \includegraphics[width=0.99\columnwidth]{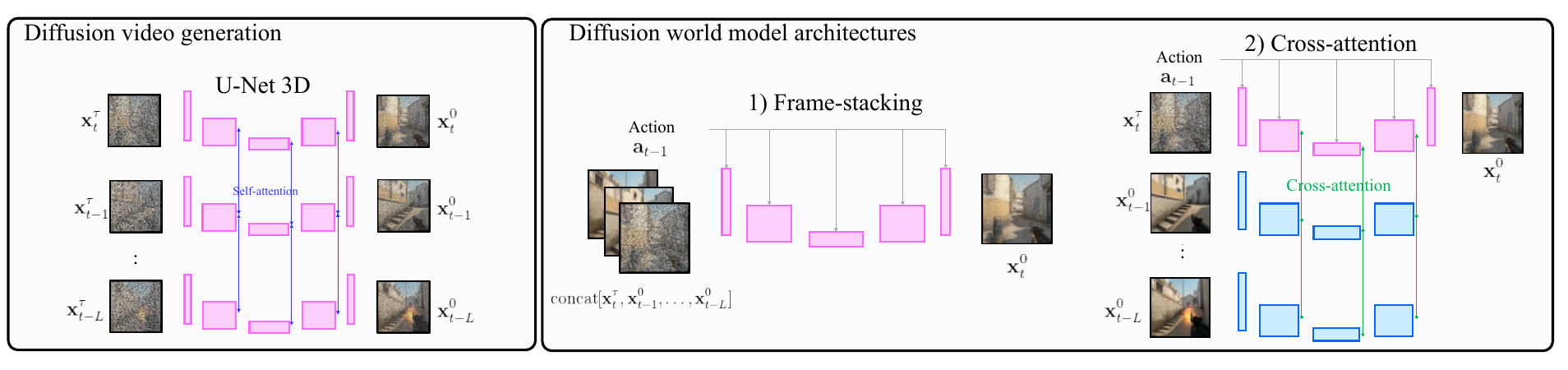}
    % \vskip -0.18in
    \caption{We tested two architectures for \textsc{diamond}'s diffusion model which condition on previous image observations in different ways. To illustrate differences with typical video generation models, we also visualize a U-Net 3D \citep{unet3d} which diffuses a block of frames simultaneously.}
    \label{fig_architectures}
    \end{center}
\end{figure}

We consider two potential diffusion model architectures, summarized in Figure \ref{fig_architectures}.

\textbf{Frame-stacking.} The simplest way to condition on previous observations is by concatenating the previous $L$ frames together with the next noised frame, $\operatorname{concat}[ \x_t^{\tau}, \x_{t-1}^0, \dots, \x_{t-L}^0]$, which is compatible with a standard U-Net 2D \citep{ronneberger2015unet}.
This architecture is particularly attractive due to its lightweight construction, requiring minimal additional parameters and compute compared to typical image diffusion. This is the architecture we used for the main body of the paper.
% Experiments in Section \ref{sec_experiments} show this to be a surprisingly powerful mechanism for modeling both simple and complex environments. 

\textbf{Cross-attention.} 
The U-Net 3D \citep{unet3d}, also displayed for comparison in Figure \ref{fig_architectures}, is a leading architecture in video diffusion \citep{ho2022video}. We adapted this design to have an autoregressive cross-attention architecture, formed of a core U-Net 2D, that only receives a single noised frame as direct input, but which cross-attends to the activations of a separate history encoder network. This encoder is a lightweight version of the U-Net 2D architecture. Parameters are shared for all $L$ encoders, and each receives the relative environment timestep embedding as input.
The final design differs from the U-Net 3D which diffuses all frames jointly, shares parameters across networks, and uses self-, rather than cross-, attention.

\subsection{Metrics, Baselines and Compute}
\textbf{Metrics.}
To evaluate the visual quality of generated trajectories, we use the standard Fréchet Video Distance (\textbf{FVD}) \citep{unterthiner2018towards} as implemented by \citet{skorokhodov2022stylegan}. This is computed between 1024 real videos (taken from the test set), and 1024 generated videos, each 16 frames long (1-2 seconds). Models condition on $L=6$ previous real frames, and the real action sequence. On this same data, we also report the Fréchet Inception Distance (\textbf{FID}) \citep{heusel2017gans}, which measures the visual quality of individual observations, ignoring the temporal dimension. For these same sets of videos, we also compute the \textbf{LPIPS} loss \citep{zhang2018lpips} between each \textit{pair} of real/generated observations \citep{yan2023teco}.
\textbf{Sampling rate} describes the number of observations that can be generated, in sequence, by a single Nvidia RTX A6000 GPU, per second.

\textbf{Baselines.}
We compare against two well-established world model methods; DreamerV3 \citep{hafner2023dreamerv3} and \textsc{iris} \citep{iris2023}, adapting the original implementations to train on a static dataset. We ensured baselines used a similar number of parameters to \textsc{diamond}. Two variants of \textsc{iris} are reported; image observations are discretized into $K=16$ tokens (as used in the original work), or into $K=64$ tokens (achieved with one less down/up-sampling layer in the autoencoder, see Appendix E of \citet{iris2023}), which provide the potential for modeling higher-fidelity visuals. 
% The full list of hyperparameters is provided in the Appendix.
% IRIS: 35M for autoencoder, 88M for transformer, seq length 8...

\textbf{Compute.}
All models (baselines and \textsc{diamond}) were trained for 120k updates with a batch size of 64, on up to 4$\times$A6000 GPUs. Each training run took between 1-2 days.

\subsection{Analysis}

\begin{table}[h]
\caption{Results for 3D environments. These metrics compare observations from real trajectories and generated trajectories. The generated trajectories are conditioned on an initial set of $L=6$ observations and a real sequence of actions.}
\label{tab:video_results}
\begin{center}
\resizebox{0.999 \columnwidth}{!}{
\begin{tabular}{ l c c c c c c c c}
\toprule
\multicolumn{1}{c}{}  & \multicolumn{3}{c}{ ------------ \textbf{CS:GO} ------------} & \multicolumn{3}{c}{ ----------- \textbf{Driving} ----------- } & \multicolumn{1}{c}{\bf Sample rate} & \multicolumn{1}{c}{\bf Parameters} \\
\multicolumn{1}{l}{\bf Method}  & \multicolumn{1}{c}{\bf FID $\downarrow$} & \multicolumn{1}{c}{\bf FVD $\downarrow$ } & \multicolumn{1}{c}{\bf LPIPS $\downarrow$ } & \multicolumn{1}{c}{\bf FID $\downarrow$} & \multicolumn{1}{c}{\bf FVD $\downarrow$} & \multicolumn{1}{c}{\bf LPIPS $\downarrow$ }  & \multicolumn{1}{c}{\bf (Hz) $\uparrow$} & \multicolumn{1}{c}{\bf (\#)} \\ 
\hline \\
DreamerV3 & 106.8 & 509.1 & 0.173 & 167.5 & 733.7 & 0.160 & 266.7 & 181M \\
IRIS ($K=16$) & 24.5 & 110.1 & 0.129 & 51.4 & 368.7 & 0.188 & 4.2 & 123M \\
IRIS ($K=64$) & 22.8 & 85.7 & 0.116 & 44.3 & 276.9 & 0.148 & 1.5 & 111M \\
$\textsc{diamond}$ frame-stack (ours) & 9.6 & 34.8 & 0.107 & 16.7 & 80.3 & 0.058 & 7.4 & 122M \\
$\textsc{diamond}$ cross-attention (ours) & 11.6 & 81.4 & 0.125 & 35.2 & 299.9 & 0.119 & 2.5 & 184M \\
\bottomrule
\end{tabular}
}
\end{center}
\end{table}

Table \ref{tab:video_results} reports metrics on the visual quality of generated trajectories, along with sampling rates and number of parameters, for the frame-stack and cross-attention \textsc{diamond} architectures, compared to baseline methods. 
\textsc{diamond} outperforms the baselines across all visual quality metrics. 
This validates the results seen in the wider video generation literature, where diffusion models currently lead, as discussed in Section \ref{sec:related_work}.
The simpler frame-stacking architecture performs better than cross-attention, something surprising given the prevalence of cross-attention in the video generation literature. We believe the inductive bias provided by directly feeding in the input, frame-wise, may be well suited to autoregressive generation. Overall, these results indicate \textsc{diamond} frame-stack $>$ \textsc{diamond} cross-attention $\approx$ IRIS 64 $>$ IRIS 16 $>$ DreamerV3, which we found corresponds to our intuition from visual inspection.

In terms of sampling rate, \textsc{diamond} frame-stack (with 20 denoising steps) is faster than \textsc{iris} ($K=16$). \textsc{iris} suffers from a further 2.8$\times$ slow down for the $K=64$ version, verifying its sample time is bottlenecked by the number of tokens $K$.
On the other hand, DreamerV3 is an order of magnitude faster -- this derives from its independent, rather than joint, sampling procedure, and the flip-side of this is the low visual quality of its trajectories. 

\newpage

Figure \ref{fig_generation_egs} below shows selected examples of the trajectories produced by \textsc{diamond} in CS:GO and motorway driving.
% Trajectories produced by baselines are given in Appendix \tp{x}.
% , and comparisons with baselines in Appendix Figures \tp{x} \& \tp{y}.
The trajectories are plausible, often even at time horizons of reasonable length. In CS:GO, the model accurately generates the correct geometry of the level as it passes through the doorway into a new area of the map. In motorway driving, a car is plausibly imagined overtaking on the left.

\begin{figure}[h!]
    \begin{center}
    \includegraphics[width=0.8\columnwidth, height=0.4\columnwidth]{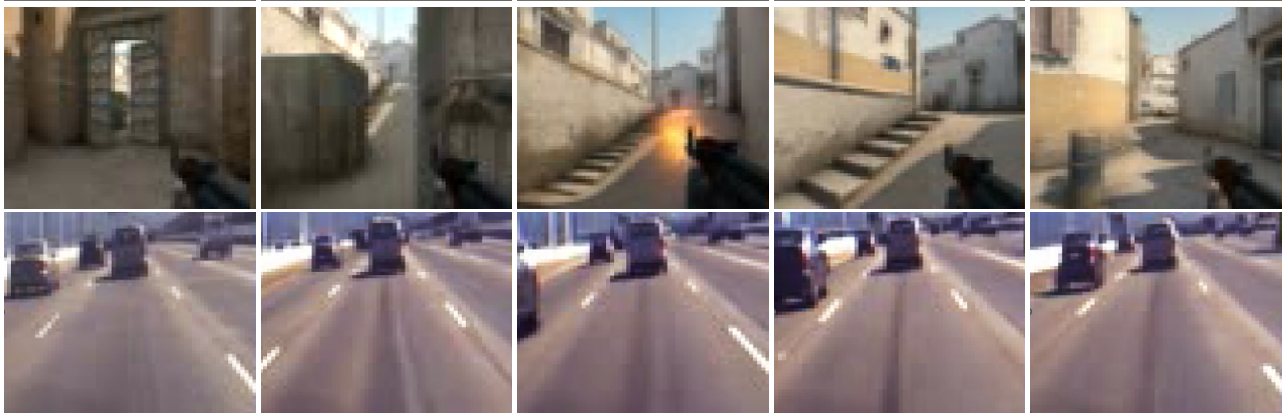}
    % \vskip -0.18in
    \caption{Example trajectories sampled every 25 timesteps from \textsc{diamond} (frame stack) for the modern 3D first-person shooter CS:GO (top row), and real-world motorway driving (bottom row).}
    \label{fig_generation_egs}
    \end{center}
\end{figure}

\begin{figure}[h!]
    \begin{center}
    \includegraphics[width=0.99\columnwidth]{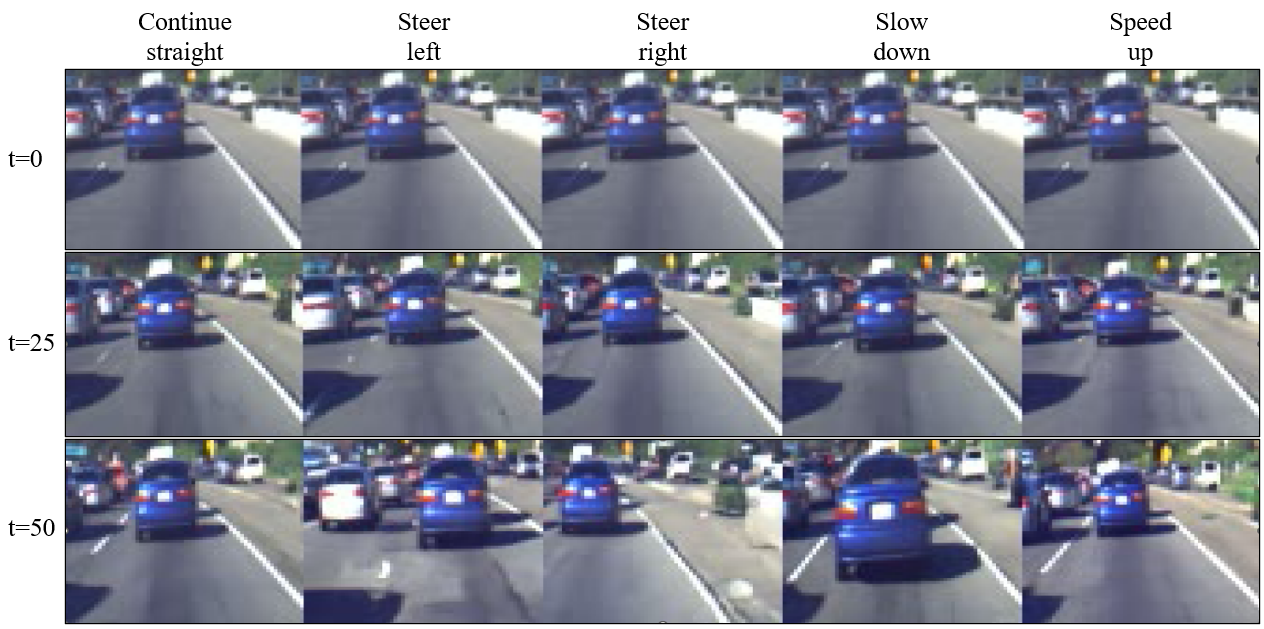}
    % \vskip -0.18in
    \caption{Effect of fixed actions on sampled trajectories in motorway driving. Conditioned on the same initial observations, we rollout the model applying differing actions. Interestingly, the model has learnt to associate 'Slow down' and 'Speed up' actions to the whole traffic slowing down and speeding up.}
    \label{fig_driving_action}
    \end{center}
\end{figure}

\begin{figure}[t!]
    \begin{center}
    \includegraphics[width=0.9\columnwidth]{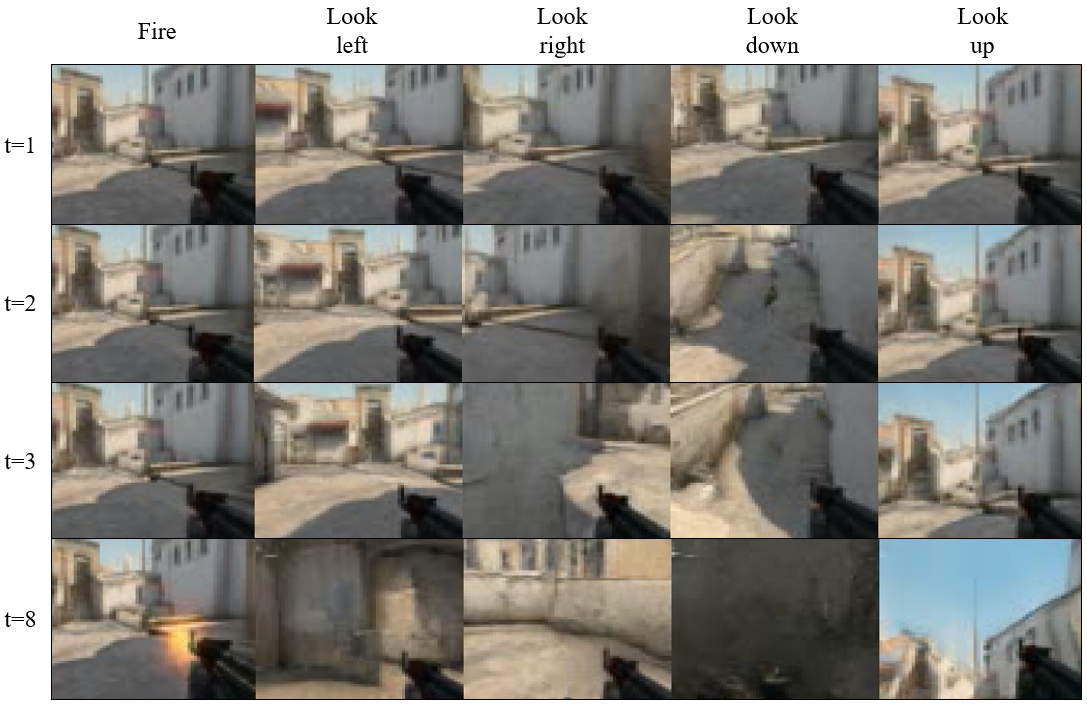}
    % \vskip -0.18in
    \caption{Effect of fixed actions on sampled trajectories in CS:GO. Conditioned on the same initial observation, we rollout the model applying differing actions. Whilst in immediate frames these have the intended effect, for longer roll-outs the observations can degenerate. For instance, it would have been very unlikely for the human demonstrator to look directly into ground in this game state, so the world model is unable to generate a plausible trajectory here, and instead snaps onto another area of the map when looking down does make sense.}
    \label{fig_csgo_action}
    \end{center}
\end{figure}

While the above experiments use real sequences of actions from the dataset, we also investigated how robust $\textsc{diamond}$ (frame stack) was to novel, user-input actions.
Figure \ref{fig_driving_action} shows the effect of the actions in motorway driving -- conditioned on the same $L=6$ real frames, we generate trajectories conditioned on five different action sequences. In general the effects are as intended, e.g. steer straight/left/right moves the camera as expected.
Interestingly, when `slow down' is input, the distance to the car in front decreases since the model predicts that the traffic ahead has come to a standstill.
%This is an interesting causal confusion, since from the action alone, slowing down should in principle \textit{increase} the distance to the car in front. 
Figure \ref{fig_csgo_action} shows similar sequences for CS:GO. For the common actions (mouse movements and fire), the effects are as expected, though they are unstable beyond a few frames, since such a sequence of actions is unlikely to have been seen in the demonstration dataset.
% We found that $\textsc{wm}$ had not learned to model the impact of less common actions such as `jump'.
We note that these issues -- the causal confusion and instabilities -- are a symptom of training world models on offline data, rather than being an inherent weakness of $\textsc{diamond}$.
% Also acknowledge the offdistribution problem, but more a problem of data than model.

\clearpage

%%%%%%%%%%%%%%%%%%%%%%%%%%%%%%%%%%%%%%%%%%%%%%%%%%%%%%%%%%%%

\end{document}